\RequirePackage{fix-cm}
\documentclass[letterpaper]{article}
\usepackage[preprint]{aaai2027}
\usepackage[hyphens]{url}
\usepackage{graphicx}
\usepackage{natbib}
\usepackage{caption}
\usepackage{booktabs}
\usepackage{multirow}
\usepackage{makecell}
\usepackage{amsmath}
\usepackage{amssymb}

\renewcommand{\topfraction}{0.95}
\renewcommand{\dbltopfraction}{0.95}
\renewcommand{\bottomfraction}{0.6}
\renewcommand{\textfraction}{0.05}
\renewcommand{\floatpagefraction}{0.9}
\renewcommand{\dblfloatpagefraction}{0.9}

\title{R3D: Revisiting 3D Policy Learning}
\author{
    \fontsize{11}{14}\selectfont\normalfont
    Zhengdong Hong\textsuperscript{*1},
    Shenrui Wu\textsuperscript{*2},
    Haozhe Cui\textsuperscript{*2},
    Boyi Zhao\textsuperscript{2},
    Ran Ji\textsuperscript{2},
    Yiyang He\textsuperscript{2},\\
    Hangxing Zhang\textsuperscript{2},
    Zundong Ke\textsuperscript{2},
    Jun Wang\textsuperscript{1},
    Guofeng Zhang\textsuperscript{1}, and
    Jiayuan Gu\textsuperscript{$\dagger$2}
}
\affiliations{
    \textsuperscript{1}Zhejiang University\qquad
    \textsuperscript{2}ShanghaiTech University
}

\begin{document}
\maketitle
\thispagestyle{plain}
\begingroup
\renewcommand{\thefootnote}{\ensuremath{\star}}
\footnotetext[1]{Equal contribution.\quad\textsuperscript{$\dagger$}Corresponding author.}
\endgroup
\setcounter{footnote}{0}

\begin{abstract}
3D policy learning offers a promising path toward improved generalization and cross-embodiment transfer, yet its progress has been hindered by underexamined design choices that have been widely inherited across methods. We systematically revisit these choices and uncover two overlooked issues. First, we identify a problematic implementation in a widely used codebase that impedes the scaling of 3D policies. Second, we show that the absence of 3D data augmentation is a critical source of unstable training and overfitting. Building on these findings, we develop a recipe that combines a transformer-based 3D encoder, a diffusion decoder, and 3D pretraining. Through extensive studies of normalization, augmentation, model scaling, pretraining, and multi-camera fusion, our approach substantially outperforms strong 3D baselines on challenging simulated and real-world manipulation tasks, providing a solid foundation for scalable 3D imitation learning.
\end{abstract}

\section{Introduction}
\label{sec:intro}
Imitation learning has been significantly shaped by 2D-centric policies, notably Diffusion Policy \citep{chi2024diffusionpolicyvisuomotorpolicy} and ACT \citep{zhao2023learning}, whose paradigms have been scaled up in modern Vision-Language-Action (VLA) models \citep{zitkovich2023rt,kim2024openvla,black2024pi_0,bu2025univla}.
However, relying on 2D projections of an inherently 3D world hinders generalization to novel viewpoints \citep{xie2024gengap} and complicates cross-embodiment transfer \citep{chen2024mirage}. 3D policy learning \citep{ze20243d,ke20243ddiffuseractorpolicy,ze2024idp3} is emerging as a compelling alternative: by reasoning over a 3D representation, such policies reduce sensitivity to camera configuration, and provide a shared spatial reference frame for transferring skills across robot embodiments.

Despite its promise, progress in 3D policies has been limited by counterintuitive observations. DP3 \citep{ze20243d} identified a ``scaling paradox'' where a lightweight PointNet backbone outperformed more powerful architectures, and 3D policy training is known to be unstable, especially at larger capacity. These two issues---\textbf{unexpected scaling behavior and unstable training}---severely hinder the adoption of modern 3D encoders \citep{wu2024pointtransformerv3,zhou2023uni3d,zhou2024point}. Several other design assumptions also remain insufficiently examined. For example, DP3 reports that Layer Normalization significantly outperforms Batch Normalization (BN) in 3D policies; ManiFlow \citep{yan2025maniflowgeneralrobotmanipulation} scaled the decoder substantially while keeping a PointNet encoder; and most 3D policies use a single camera even though point clouds naturally fuse multiple views. Rather than introducing another architectural component, \textbf{we revisit these design choices through controlled and systematic ablations}, following a line of empirical inquiry that has repeatedly advanced understanding in deep learning \citep{zhang2017understanding}. 

Our analysis reveals two crucial findings. First, we trace the reported failure of BN in 3D policies to \textbf{a problematic implementation in the Exponential Moving Average (EMA) } of a widely used codebase~\cite{diffusionpolicygithub}. Specifically, the EMA model does not correctly update BN statistics, resulting in degraded performance at inference. Consequently, subsequent methods built upon this codebase, including DP3, may inherit the same issue. After correcting this issue, we find that different normalization methods perform comparably. This result revises DP3's conclusion that BN should be avoided and removes a long-standing barrier to adopting modern pretrained 3D encoders. Second, we observe that \textbf{data augmentation} is absent from quite a few existing 3D policy implementations, despite its established role in stabilizing training and mitigating overfitting.

Building on these corrected foundations, we derive practical design principles for scaling 3D policy learning. Our architecture combines a high-capacity \textbf{transformer-based 3D encoder}, inspired by recent advances in 3D segmentation \citep{zhou2024point}, with a Diffusion Transformer decoder \citep{peebles2023scalable} that \textbf{preserves the spatial resolution of 3D features} instead of compressing them into a single global representation. This design better exploits large-capacity 3D encoders, supports \textbf{pretraining} on 3D segmentation tasks, and accommodates an \textbf{auxiliary prediction objective} that jointly decodes end-effector poses and joint configurations to strengthen proprioceptive grounding.

Concretely, our revisiting study addresses five questions through systematic experimentation:\\
\textbf{Q1:} Is BN really unsuitable for 3D policies? \emph{No---the failures stem from an EMA bug; once fixed, different normalization schemes perform comparably.}\\
\textbf{Q2:} Which 3D augmentations matter? \emph{Every technique helps, and their combination is the best, especially on hard tasks.}\\
\textbf{Q3:} Do stronger, pre-trained 3D encoders pay off? \emph{Yes---with the right pre-training task, and the optimal capacity grows with point cloud density.}\\
\textbf{Q4:} Is a deeper action decoder better? \emph{No---moderate depth suffices with fast enough speed; excessive depth overfits.}\\
\textbf{Q5:} Does multi-camera fusion still help once policies are 3D? \emph{Yes---fused multi-view point clouds substantially improve real-world performance.}\\
Answering these questions, our model \textbf{significantly outperforms} strong 3D imitation learning baselines on simulated benchmarks (RoboTwin~2.0, ManiSkill2, and Meta-World) and real-world tasks, providing extensive empirical references for future 3D policies.

\section{Related Work}
\label{sec:related}

\noindent\textbf{2D policy learning.}
Early visuomotor policies rely on 2D RGB inputs: diffusion-based approaches \citep{chi2024diffusionpolicyvisuomotorpolicy,liu2025rdt1bdiffusionfoundationmodel,yu2025efficienttaskspecificconditionaldiffusion} achieve stable behavior cloning via generative modeling, while transformer architectures \citep{zhao2023learning,shridhar2022perceiveractormultitasktransformerrobotic} leverage spatiotemporal attention for long-horizon manipulation. Recent 2D VLAs \citep{kim2024openvla,kim2025fine,black2024pi_0,zitkovich2023rt,o2024open,li2023vision,team2024octo,wu2023unleashing,bu2025univla} couple large vision--language backbones with action policies. Lacking explicit depth, these image-space models show weak geometric reasoning, poor viewpoint generalization, and reduced robustness in clutter.

\noindent\textbf{Policy learning with implicit 3D.}
A line of work instead derives implicit or lifted geometry from RGB, via multi-view fusion \citep{qian2025gp33dgeometryawarepolicy}, semantic--geometric cues \citep{ni2025vodpsemanticgeometricadaptivediffusion}, lifting 2D pretrained models into 3D-aligned feature spaces \citep{jia2024lift3dfoundationpolicylifting}, or multi-view pretraining \citep{qian20253dmvp3dmultiviewpretraining,Hou_2025_ICCV}; VLA frameworks \citep{yuan2025depthvlaenhancingvisionlanguageactionmodels,qu2025spatialvlaexploringspatialrepresentations,bhat20253d,li2025spatialforcingimplicitspatial} similarly inject depth or spatial encodings. This 2.5D regime mitigates projection artifacts but remains limited by partial 3D awareness and weak spatial consistency between perception and action.

\noindent\textbf{3D policy learning.}
Explicit 3D policies span diffusion-based formulations \citep{ze20243d,ze2024idp3,ke20243ddiffuseractorpolicy,chen2025acditadaptivecoordinationdiffusion,zhao2025anchordp33daffordanceguided}, transformer-based architectures \citep{yang2025fp33dfoundationpolicy,gervet2023act3d3dfeaturefield,pmlr-v229-goyal23a}, 3D-aware VLAs \citep{zhen20243dvla3dvisionlanguageactiongenerative,sun2025geovlaempowering3drepresentations,li2025pointvlainjecting3dworld,li2025bridgevlainputoutputalignmentefficient}, and flow-based methods \citep{gkanatsios20253dflowmatchactorunified,yan2025maniflowgeneralrobotmanipulation}. Yet they still face training instability, the cost of dense 3D encoding, and redundant representations---motivating our focus on a lean, scalably trained transformer-based 3D diffusion formulation.

\section{Revisiting 3D Policy Learning}
\label{sec:revisiting}

We first investigate why prior methods fail to benefit from more powerful 3D backbones. The cause turns out to be not architectural novelty but subtle yet decisive implementation details, which we distill into actionable training principles. Throughout this section we use RoboTwin 2.0 \citep{chen2025robotwin} as the testbed, reporting success rates on representative bimanual tasks (settings in Appendix~\ref{app:simulation}).

\subsection{Resolving the 3D Policy Scaling Paradox}
\label{sec:bn_ln}

Prior work on 3D policies has presented two seemingly independent but highly consequential observations. First, DP3 \citep{ze20243d} reported a ``scaling paradox'' where modern, high-capacity architectures (e.g., PointNet++, PointNeXt, and Point Transformer) suffer catastrophic performance drops compared to lightweight PointNet backbones. Second, DP3's ablation observed that Batch Normalization (BN) destabilizes training, concluding that LayerNorm (LN) must be used instead. A critical confounding variable was overlooked: the advanced backbones in these scaling experiments all use BN by default, so the failure of BN was conflated with an inability to scale capacity. Since scaling up the 3D encoder means adopting large-scale pre-trained models, which predominantly rely on BN, clearing this barrier is a prerequisite for our study.

The root cause of both phenomena is neither model capacity nor the BN algorithm itself, but a bug in the widely used Diffusion Policy codebase \citep{chi2024diffusionpolicyvisuomotorpolicy}: the framework maintains an Exponential Moving Average (EMA) copy of the policy for inference but \textbf{fails to synchronize BN buffers} (running mean and variance). Unlike LN or RMSNorm \citep{zhang2019root}, which compute statistics per-instance and carry no state, BN relies on these buffers; with the buggy EMA the rollout network normalizes activations with mismatched statistics, driving the robot into out-of-distribution states where errors compound over an episode. Every BN-based backbone evaluated in DP3 silently inherited this bug, creating the dual illusion that high-capacity 3D backbones cannot be optimized and that BN is unusable in policies.

\noindent\textbf{Re-evaluating Normalization and Model Scaling.}
After fixing the EMA implementation, the scaling paradox is resolved. Table~\ref{tab:norm_compare} shows that BN-based policies now perform competitively with LN. Specifically, DP3 with PointSAM achieves the highest success rates across both normalization methods, proving the transferability of pre-trained backbones to policy learning. We highlight this codebase flaw to prevent future propagation. Consequently, we use LN in subsequent experiments to align with our pre-trained transformer backbones, rather than out of necessity.
\begin{table}[t]
\centering
\small
\setlength{\tabcolsep}{3pt}
\begin{tabular}{l ccc c}
\toprule
Method & \makecell{Beat\\Hammer} & \makecell{Move\\Card} & \makecell{Place\\Shoe} & \textbf{Avg.} \\
\midrule
\multicolumn{5}{l}{\emph{w/ EMA bug (original Diffusion Policy codebase)}} \\
DP3 (PointNet+BN) & 0 & 3 & 0 & 1.0 \\
DP3 (Uni3D+BN)    & 0 & 0 & 0 & 0.0 \\
\midrule
\multicolumn{5}{l}{\emph{w/ fixed EMA (ours)}} \\
DP3 (PointNet+BN) & 70 & 56 & 40 & 55.3 \\
DP3 (PointNet+LN) & 72 & 60 & 41 & 57.6 \\
DP3 (Uni3D+BN)    & 84 & 60 & 42 & 62.0 \\
DP3 (Uni3D+LN)    & 86 & 60 & 48 & 64.7 \\
DP3 (PointSAM+RMSNorm) & 90 & 52 & 48 & 63.3 \\
DP3 (PointSAM+BN)  & 94 & 60 & 40 & 64.7 \\
DP3 (PointSAM+LN)      & 90 & 58 & 60 & \textbf{69.3} \\
\bottomrule
\end{tabular}
\caption{\textbf{Effect of normalization under different backbones.} With the buggy EMA inherited from the original Diffusion Policy implementation, BN-based policies collapse. After fixing the EMA implementation, BN performs comparably to LN. For PointSAM, LayerNorm and RMSNorm also achieve comparable performance. Results are reported on the three common RoboTwin 2.0 tasks (\% success rate).}
\label{tab:norm_compare}
\end{table}

\subsection{Mitigating Overfitting via Data Augmentation}
\label{sec:fix_overfitting}
3D policy training is notoriously vulnerable to overfitting \citep{ze20243d}, often compelling researchers to report peak performance over a range of checkpoints rather than final convergence. Data augmentation is the standard remedy, yet existing 3D policies use it sparsely and inconsistently: DP3 \citep{ze20243d} applies no augmentation at all; FP3 \citep{yang2025fp33dfoundationpolicy} reports only point cloud dropout; and ManiFlow \citep{yan2025maniflowgeneralrobotmanipulation} adopts a limited subset while explicitly arguing against certain 3D augmentations---an argument our experiments below do not support. 
We revisit this design space systematically: we assemble a pipeline of common techniques, ablate each one individually, and identify the combination that works best.
\textbf{1.~FPS Randomization:} following PointNet++ \citep{qi2017pointnet++}, we randomize the point order fed to FPS at each step, preventing reliance on the sampler's deterministic order. \textbf{2.~Point Cloud Color Jitter:} inspired by QT-Opt \citep{kalashnikov2018qtoptscalabledeepreinforcement}, we jitter the RGB channels' brightness, contrast, and saturation. \textbf{3.~Additive Noise and Dropout:} we add modest Gaussian noise to point coordinates and proprioceptive vectors, and randomly drop points, forcing accurate prediction from incomplete scenes. Exact ranges are in Appendix~\ref{app:augmentation}.

Our training and validation curves (see Appendix~\ref{app:augmentation}) evidence the impact: without augmentation, learning curves fluctuate heavily and success rates decline as training progresses; with augmentation, training stabilizes and success rates improve substantially thanks to broader out-of-distribution exposure.

To disentangle the contribution of each technique, Table~\ref{tab:aug_ablation} applies one augmentation at a time on three ``Easy'' and three ``Hard'' RoboTwin 2.0 tasks. Every individual augmentation helps, data augmentation is especially crucial on ``Hard'' tasks with heavy distractors, omitting augmentation entirely (DP3 setting) yields the worst results, and combining all techniques is the best overall. For a direct comparison with prior practice, we also evaluate the exact augmentation combinations used by ManiFlow and FP3 on our pipeline; both underperform our full combination. 

\begin{table}[t]
\centering
\small
\setlength{\tabcolsep}{2pt}
\resizebox{\columnwidth}{!}{%
\begin{tabular}{l ccc ccc}
\toprule
& \multicolumn{3}{c}{``Easy'' Tasks} & \multicolumn{3}{c}{``Hard'' Tasks} \\
\cmidrule(lr){2-4} \cmidrule(lr){5-7}
Data Aug.\ Type & \makecell{Beat\\Hammer} & \makecell{Move\\Card} & \makecell{Place\\Shoe} & \makecell{Beat\\Hammer} & \makecell{Lift\\Pot} & \makecell{Burger\\Fries} \\
\midrule
No Aug.\ (as in DP3) & 83 & 67 & 79 & 15 & 44 & 78 \\
Color Jitter      & 87 & \underline{79} & \underline{86} & 27 & 47 & \underline{92} \\
PCD Dropout       & 89 & 75 & 81 & \underline{29} & \underline{62} & 83 \\
FPS Randomization & \underline{90} & 72 & 83 & 22 & 54 & 79 \\
Robot State Noise & 86 & 76 & 83 & 23 & 57 & 82 \\
PCD Noise         & 86 & 74 & 83 & 20 & 54 & 85 \\
ManiFlow's Aug.\ Combo & 85 & 74 & 82 & 18 &  49& 82 \\
FP3's Aug.\ Combo      & 87 & 75 & 85 & 26 &  55& 90 \\
All Aug.\ (Ours)  & \textbf{91} & \textbf{85} & \textbf{87} & \textbf{37} & \textbf{67} & \textbf{95} \\
\bottomrule
\end{tabular}%
}
\caption{Ablation of individual data augmentation techniques (\%) on RoboTwin 2.0. Each middle row applies a single technique. \textbf{Bold}: best; \underline{underline}: second-best per column.}
\label{tab:aug_ablation}
\end{table}

\section{Revisiting 3D Policy Architecture}
\label{sec:architecture}

\subsection{Unifying 3D Perception and 3D Policies}

\begin{figure*}[t]
\centering
\includegraphics[width=0.8\textwidth]{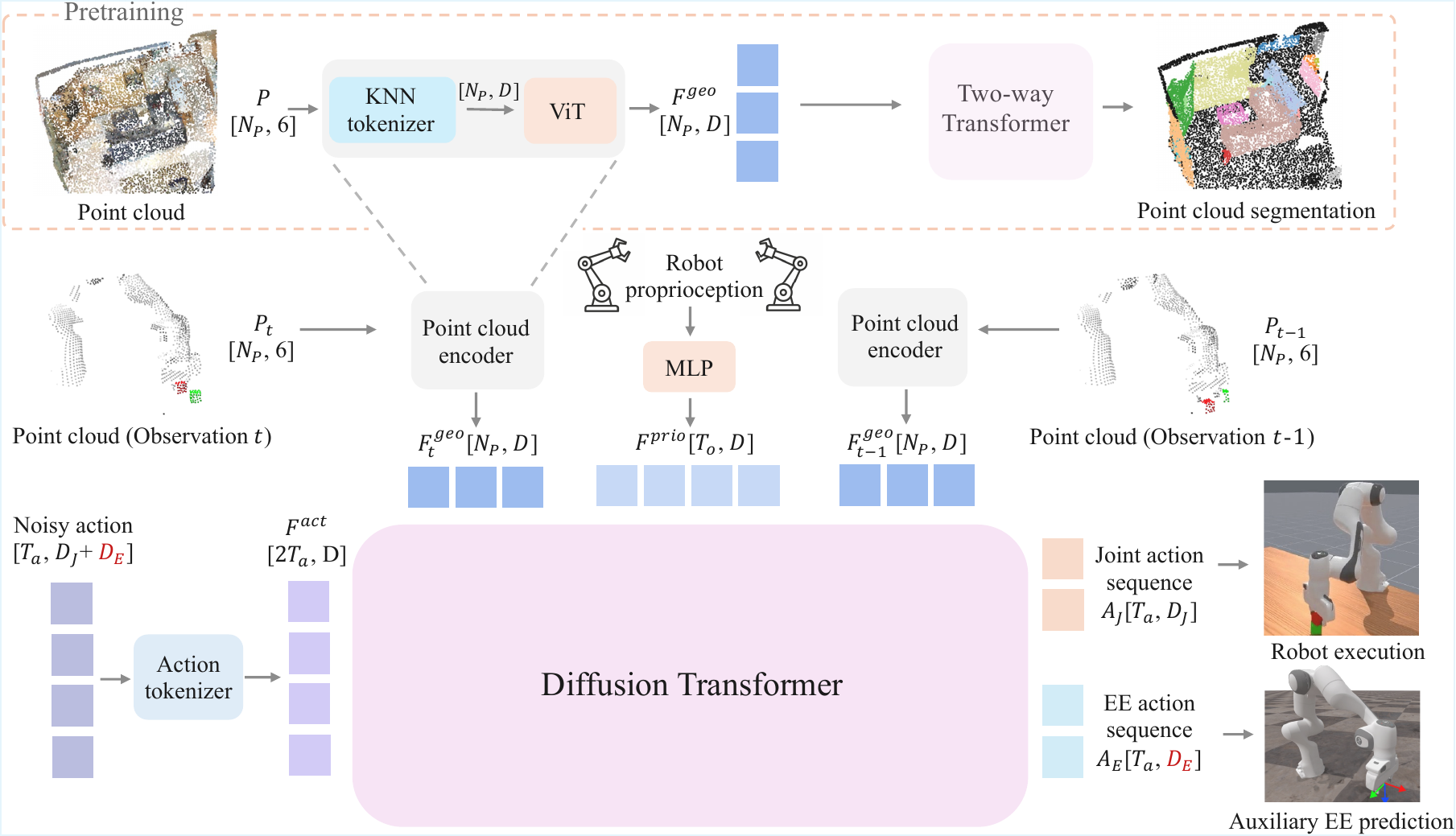}
\caption{\textbf{Pipeline architecture}: a point cloud encoder encodes the input point clouds and a diffusion transformer decodes actions. The encoder can be pre-trained on 3D segmentation tasks.}
\label{fig:pipe-overall}
\end{figure*}

3D perception and 3D policy learning share a core mechanism: interaction between geometric features and task-specific queries. Point-SAM \citep{zhou2024point} cross-attends point cloud tokens with prompt tokens, while diffusion policies \citep{chi2024diffusionpolicyvisuomotorpolicy} cross-attend observation tokens with noisy action queries. Our architecture (Figure~\ref{fig:pipe-overall}) is therefore not a new invention but a synthesis of established components---a transformer point cloud encoder from 3D perception and a Diffusion Transformer action decoder compatible with recent 3D policies such as ManiFlow---assembled so each design dimension can be studied in isolation: the \emph{encoder} (structure, capacity, pre-training task) and the \emph{decoder} (conditioning, depth, auxiliary objectives).

\subsection{Preliminaries}
\label{sec:preliminaries}
At each timestep $t$, the observation $O_t$ consists of a colored point cloud $P_t \in \mathbb{R}^{N_P \times 6}$ (sampled or padded to a fixed size $N_P$) and a proprioceptive state $q_t \in \mathbb{R}^{N_q}$. Given an observation horizon $T_o$, the goal is to predict an action sequence $\mathbf{a}_t = \{a_t, \dots, a_{t+T_a-1}\}$ over horizon $T_a$. Following \citet{chi2024diffusionpolicyvisuomotorpolicy}, we model the policy as a conditional diffusion process that refines a noise sequence $\mathbf{a}^{(K)} \sim \mathcal{N}(0, \mathbf{I})$ into a clean action sequence $\mathbf{a}^{(0)}$ over $K$ iterations, conditioned on $O_{t-T_o+1:t}$; a Diffusion Transformer (DiT) serves as the noise prediction network $\epsilon_\theta$.

\subsection{Revisiting the Encoder: Beyond PointNet}
\label{sec:encoder}
Prior 3D policies almost universally adopt a lightweight PointNet encoder \citep{ze20243d,yan2025maniflowgeneralrobotmanipulation}. With the normalization obstacle cleared (Section~\ref{sec:bn_ln}), we ask whether a stronger encoder pays off. We adopt a Transformer-based backbone from recent point cloud segmentation work \citep{zhou2024point}; it uses Layer Normalization throughout, matching the pre-trained transformer weights we initialize from.

The encoder partitions the input point cloud $\mathcal{P} \in \mathbb{R}^{N_P \times 3}$ into local patches---Farthest Point Sampling (FPS) selects $N_C$ center points and $k$-NN groups their neighborhoods---projects each patch into a $D$-dimensional embedding via a lightweight PointNet \citep{qi2017pointnet} with LN, adds positional embeddings of patch centers, and processes the tokens through self-attention layers initialized from a pre-trained 2D ViT \citep{dosovitskiy2020image}. Crucially, it outputs $N_C$ structured point tokens $F^{geo} \in \mathbb{R}^{N_C \times D}$ representing fine-grained geometry. This contrasts with DP3 \citep{ze20243d}, which collapses features into a global descriptor, and Uni3D \citep{zhou2023uni3d}, which appends a $[CLS]$ token. We omit global pooling, preserving the full spatial resolution of the patch tokens to give the decoder a dense geometric representation.

\subsection{Revisiting Encoder Pre-Training}
\label{sec:pretraining}
Since the encoder structure comes from 3D segmentation \citep{zhou2024point}, we can pretrain it on large-scale 3D scene datasets such as ScanNet \citep{dai2017scannet}, ARKitScenes \citep{baruch2021arkitscenes}, and PartNeXt \citep{wangpartnext}. Pre-training injects geometric priors that reduce the task-specific samples needed for fine-tuning and improve robustness to real-world shifts in sensor noise, occlusion, and clutter.

\noindent\textbf{Which 3D pre-training task transfers best?}
We choose 3D segmentation because 3D policies require spatial understanding of scenes, objects, and parts. To validate this, we compare encoders pre-trained on two alternatives---3D-image-text alignment (Uni3D \citep{zhou2023uni3d}) and 3D generation (TripoSG \citep{li2025triposg})---each against its own train-from-scratch counterpart. Table~\ref{tab:pretrain_tasks} shows efficacy varies by encoder and task: most configurations gain over scratch, but 3D generation degrades some tasks. Segmentation and alignment transfer most reliably.

\begin{table}[t]
\centering
\small
\setlength{\tabcolsep}{2.5pt}
\begin{tabular}{l cccc}
\toprule
Pretraining Config & \makecell{Move\\Card\\(easy)} & \makecell{Place\\Shoe\\(easy)} & \makecell{Beat\\Hammer\\(hard)} & \makecell{Lift\\Pot\\(hard)} \\
\midrule
3D Segmentation          & \textbf{75} & 83 & \textbf{24} & 56 \\
\ \ PointSAM (scratch)   & 66 & 50 & 19 & 51 \\
\midrule
3D-Image-Text Alignment  & 72 & \textbf{93} & 17 & \textbf{69} \\
\ \ Uni3D (scratch)      & 59 & 51 & 10 & 51 \\
\midrule
3D Generation            & 8  & 0  & 14 & 59 \\
\ \ TripoSG (scratch)    & 50 & 4  & 12 & 46 \\
\bottomrule
\end{tabular}
\caption{Comparison of pre-training tasks (\%) on RoboTwin 2.0. No data augmentation or auxiliary end-effector objective is used. Each pre-trained encoder is paired with its train-from-scratch counterpart.}
\label{tab:pretrain_tasks}
\end{table}

\subsection{Action Decoding with Dense Conditioning}
\label{sec:decoder}

The action decoder is a Diffusion Transformer \citep{peebles2023scalable} operating on noisy actions $\mathbf{a}^{(k)}_t = (a^{(k)}_t, \dots, a^{(k)}_{t+T_a-1})$ at diffusion iteration $k$. Rather than conditioning on a global latent vector, it cross-attends the action queries directly to the full set of geometric tokens. This dense conditioning is shared by recent 3D policies such as ManiFlow \citep{yan2025maniflowgeneralrobotmanipulation} and FP3 \citep{yang2025fp33dfoundationpolicy}; we claim it not as a contribution but verify it against global-feature conditioning in our ablations.

The decoder processes three token types in a shared $D$-dimensional space: \textbf{geometric tokens} $\mathcal{F}^{geo} \in \mathbb{R}^{(T_o \cdot N_P) \times D}$, the point cloud tokens over the observation horizon; \textbf{proprioception tokens} $\mathcal{F}^{prio} \in \mathbb{R}^{T_o \times D}$, from an MLP on $q_t$; and \textbf{action tokens} $\mathcal{F}^{act} \in \mathbb{R}^{T_a \times D}$, from an MLP on $a^{(k)}_t$. Learnable temporal embeddings encode each token's relative timestep, and a sinusoidal embedding of $k$ is prepended to the action sequence. The DiT takes the observation tokens as cross-attention context and denoises the action tokens into chunks.

The benefit is \textbf{preserved spatial resolution}: avoiding the global-pooling bottleneck lets the DiT attend to task-relevant regions---a handle, a container edge---during denoising, which matters most for fine-grained spatial reasoning (schematic in Appendix~\ref{app:dense_conditioning}).

\subsection{Multi-Objective Decoding for Proprioceptive Grounding}
\label{sec:auxiliary}
To further regularize learning and improve control accuracy, we study an auxiliary task that decodes task-space actions (end-effector poses) alongside joint-space actions. Since the end-effector pose is linked to the joint configuration by forward kinematics, this objective adds supervision at no extra data cost.

Specifically, we include an auxiliary set of action tokens $^{ee}\mathcal{F}^{act} \in \mathbb{R}^{T_a \times D}$, embedded from noisy 7D task-space actions (3D translation and 4D quaternion in the base frame), and denoise joint-space and task-space tokens simultaneously. A \emph{causal attention mask} in the decoder's self-attention layers lets task-space tokens attend to joint-space tokens but not vice versa, so the auxiliary task provides features for proprioceptive grounding without interfering with the primary joint-control objective. This multi-objective optimization grounds the robot's proprioception within the 3D scene, leading to better task-space alignment.

\subsection{Revisiting Network Scaling}
\label{sec:scaling}

Prior 3D policies offer little guidance on how to size the network: DP3 implies smaller is better, while ManiFlow fixes a 12-layer decoder across all benchmarks without ablation. We systematically analyze the impact of encoder size and decoder depth on policy performance.

\begin{table}[t]
\centering
\small
\setlength{\tabcolsep}{2.5pt}
\begin{tabular}{l ccc}
\toprule
Setting & \makecell{tiny\\(53MB)} & \makecell{small\\(115MB)} & \makecell{base\\(366MB)} \\
\midrule
1024 pts, 50 demos  & \textbf{83.8} & 76.8 & 73.0 \\
1024 pts, 100 demos & \textbf{95.3} & 89.0 & 86.5 \\
1024 pts, 200 demos & \textbf{97.8} & 92.4 & 88.2 \\
8192 pts, 50 demos  & 68.0 & \textbf{81.8} & 80.2 \\
\bottomrule
\end{tabular}
\caption{Encoder size vs.\ point cloud resolution and number of demonstrations, averaged over 5 RoboTwin 2.0 ``Easy'' tasks (\%). The optimal encoder size grows with input point cloud density and is stable across demonstration counts; per-task results are in Appendix~\ref{app:encoder_scaling}.}
\label{tab:encoder_size}
\end{table}

\begin{table}[t]
\centering
\small
\setlength{\tabcolsep}{2pt}
\begin{tabular}{llcccccc}
\toprule
\textbf{Task} & \textbf{Encoder}
& \textbf{2} & \textbf{4} & \textbf{6} & \textbf{8} & \textbf{10} & \textbf{12} \\
\midrule
\multirow{2}{*}{Move Card}
& Light (ManiFlow)
& \textbf{71} & 58 & 53 & 62 & 63 & 61 \\
& Heavy (Ours)
& 77 & 85 & 83 & \textbf{90} & 83 & 78 \\
\midrule
\multirow{2}{*}{Place Shoe}
& Light (ManiFlow)
& 48 & 43 & \textbf{50} & 47 & 47 & 46 \\
& Heavy (Ours)
& 83 & 87 & 88 & \textbf{90} & 90 & 85 \\
\midrule
\multirow{2}{*}{Turn Switch}
& Light (ManiFlow)
& 44 & 42 & \textbf{49} & 38 & 45 & 49 \\
& Heavy (Ours)
& 43 & 56 & 50 & \textbf{58} & 53 & 57 \\
\midrule
\multirow{2}{*}{\textbf{Average}}
& Light (ManiFlow)
& \textbf{54.3} & 47.7 & 50.7 & 49.0 & 51.7 & 52.0 \\
& Heavy (Ours)
& 67.7 & 76.0 & 73.7 & \textbf{79.3} & 75.3 & 73.3 \\
\bottomrule
\end{tabular}
\caption{Success rates (\%) vs. decoder depth (number of blocks) for a light PointNet encoder (as in ManiFlow) and our transformer encoder, with the rest of the pipeline held fixed. Twelve blocks is past the optimum for both encoders. \textbf{Bold:} highest per row.
}
\label{tab:decoder_depth_comparison}
\end{table}

\noindent\textbf{Encoder Size.} With the EMA bug fixed (Section~\ref{sec:bn_ln}), stronger transformer encoders such as Uni3D \citep{zhou2023uni3d} become trainable and outperform PointNet \citep{qi2017pointnet}. Table~\ref{tab:encoder_size} shows that the proper encoder size depends on the input density: under the standard 1024-point RoboTwin 2.0 setting, ViT-tiny (53MB) is optimal, whereas the 8192-point real-world setting favors ViT-small (115MB). That is, as point cloud density increases, the encoder should be scaled accordingly to provide sufficient representational capacity (each encoder is pre-trained with the corresponding number of points). Varying the number of demonstrations under the 1024-point setting confirms this choice is stable: performance improves with more demonstrations across all sizes, and ViT-tiny remains the best throughout (50--200 demos).

\begin{figure*}[t]
\centering
\includegraphics[width=\textwidth]{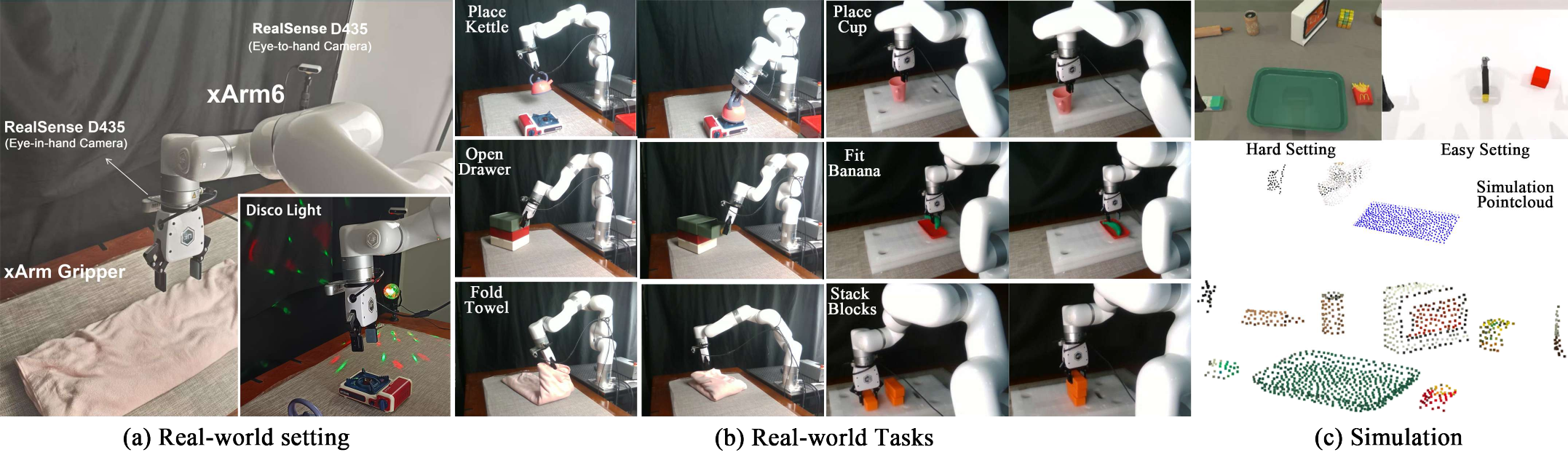}
\caption{\textbf{Experimental Setup.} Real-world setup with a disco light (in (a)) used for the policy robustness test in Table~\ref{tab:single_view}.}
\label{fig:exp-overall}
\end{figure*}

\noindent\textbf{Decoder Depth.}
While ManiFlow \citep{yan2025maniflowgeneralrobotmanipulation} employs a 12-layer DiTX decoder, we observed severe overfitting in real-world experiments: the policy often failed under significant object pose variations despite achieving low training loss. To systematically investigate this, we conducted an ablation study on decoder depth (number of blocks) using the RoboTwin 2.0 benchmark.
As shown in Table~\ref{tab:decoder_depth_comparison}, two key patterns emerge. First, the optimal decoder depth typically lies between 4 and 8 blocks; scaling beyond 8 layers leads to a noticeable performance decline, confirming that excessive capacity hampers generalization. Second, we observe a synergy between encoder capacity and decoder depth: the "heavy" R3D encoder reaches its peak performance at a greater depth (8 blocks) compared to ManiFlow's "light" encoder, which peaks much earlier (e.g., at 2 or 6 blocks). This suggests that more expressive representations require a commensurate decoder capacity to be fully utilized.

Depth also governs inference cost, as it dominates inference time. We therefore adopt \textbf{4 blocks}, sacrificing 3.3 points relative to the 8-block peak to achieve the lowest latency among all measured methods  (Table~\ref{tab:latency}; protocol in Appendix~\ref{app:latency}). Notably, despite having the highest-capacity encoder of the five models, our approach reaches an overall inference time of just 65.26\,ms. This efficiency is driven by the decoder rather than the encoder: our 4-block DiT decoder requires only 35.63\,ms, whereas DP3’s UNet takes 98.78\,ms. Moderate depth is thus favorable twice over: it prevents the overfitting typical of deeper decoders and ensures superior execution speed.

\begin{table}[t]
\centering
\small
\setlength{\tabcolsep}{3pt}
\begin{tabular}{l ccccc}
\toprule
Method & DP & $\pi_0$ & DP3 & ManiFlow & \textbf{Ours} \\
\midrule
Latency (ms) & 140.03 & 137.88 & 115.42 & 122.24 & \textbf{65.26} \\
\bottomrule
\end{tabular}
\caption{Inference latency in the real-world setting, measured identically for all methods on one RTX 3090. Our 4-block decoder lets the highest-capacity encoder run fastest.}
\label{tab:latency}
\end{table}

\section{Experiments}
\label{sec:exp}

\subsection{Simulation Experiments}

\noindent\textbf{Setup.} We evaluate on three simulation benchmarks (Figure~\ref{fig:exp-overall}c). While all three test general manipulation ability, each is a particularly demanding testbed for one capability, which is how we organize the analysis below.
\textbf{(1) RoboTwin 2.0} \citep{chen2025robotwin} covers bimanual manipulation in an ``Easy'' setting (clean scenes) and a ``Hard'' setting (random distractors, varying lighting/textures, larger pose variances); the contrast between the two isolates generalization. We curate five tasks per setting and follow the official protocol of 50 demos per task at 1024 points.
\textbf{(2) ManiSkill2} \citep{gu2023maniskill2} contributes three tasks over short and long horizons with 1000 demonstrations each at 1024 points; following Point Cloud Matters \citep{zhu2024point}, ``PegInsertionSide'' is split into Grasp, Align, and Insert, whose tight clearances make it a stringent test of fine-grained precision.
\textbf{(3) Meta-World} \citep{yu2019metaworld} supplies a multi-task setting: on its ``Hard'' suite we train one language-conditioned policy across all tasks instead of one policy per task.
Per-task ManiSkill2 results, task visualizations, and baseline implementations are in Appendix~\ref{app:simulation}.

\begin{table*}[t]
\centering
\small
\setlength{\tabcolsep}{3pt}
\begin{minipage}[t]{0.49\textwidth}
\centering
\begin{tabular}[t]{l ccccc c}
\toprule
Method & \makecell{Beat\\Hammer} & \makecell{Move\\Card} & \makecell{Turn\\Switch} & \makecell{Open\\Micro.} & \makecell{Place\\Shoe} & \textbf{Avg.} \\
\midrule
DP+eVGGT        & 17 & 4  & 29 & 69 & 9  & 25.6 \\
Spatial Forcing & 47 & 60 & 37 & 85 & 22 & 50.2 \\
DP              & 42 & 47 & 36 & 5  & 23 & 30.6 \\
DP3             & 72 & 68 & 46 & 61 & 58 & 61.0 \\
iDP3            & \textbf{92} & 69 & 52 & 57 & 43 & 62.6 \\
RDT             & 77 & 43 & 35 & 37 & 35 & 45.4 \\
ManiFlow        & \textbf{94} & 74 & 43 & 98 & 56 & 73.0 \\
\textbf{Ours}   & 91 & \textbf{85} & \textbf{56} & \textbf{100} & \textbf{87} & \textbf{83.8} \\
\bottomrule
\end{tabular}
\par\vspace{2pt}(a) RoboTwin 2.0 ``Easy'' setting.
\end{minipage}\hfill
\begin{minipage}[t]{0.49\textwidth}
\centering
\begin{tabular}[t]{l ccccc c}
\toprule
Method & \makecell{Beat\\Hammer} & \makecell{Lift\\Pot} & \makecell{Adjust\\Bottle} & \makecell{Burger\\Fries} & \makecell{Press\\Stapler} & \textbf{Avg.} \\
\midrule
ACT+eVGGT     & 7  & 2  & 20 & 10 & 14 & 10.6 \\
DP+eVGGT      & 18 & 14 & 52 & 26 & 20 & 26.0 \\
DP3           & 8  & 10 & 62 & 18 & 15 & 22.6 \\
iDP3          & 5  & 9  & 82 & 34 & 8  & 27.6 \\
RDT           & 9  & 12 & 50 & 20 & 20 & 22.2 \\
ManiFlow      & 31 & 65 & 96 & 72 & \textbf{44} & 61.6 \\
\textbf{Ours} & \textbf{37} & \textbf{67} & \textbf{98} & \textbf{95} & 27 & \textbf{64.8} \\
\bottomrule
\end{tabular}
\par\vspace{2pt}(b) RoboTwin 2.0 ``Hard'' setting.
\end{minipage}
\caption{Success rates (\%) on RoboTwin 2.0: (a) ``Easy'' and (b) ``Hard'' settings.}
\label{tab:robotwin_easy}
\end{table*}

\noindent\textbf{Results.} Our method outperforms the 2D image-based, 2.5D RGBD-based, and 3D point-cloud-based baselines on all three benchmarks, following the order introduced above.

\noindent\textbf{(1) Generalization.} Robustness in unstructured scenes is exercised most directly by the RoboTwin 2.0 ``Hard'' setting, a representative stress test combining random distractors, unseen object instances, diverse textures, variable table heights, and lighting interference. Under this randomization our method attains the highest average success rate (Table~\ref{tab:robotwin_easy}b), and it also leads on ``Easy'' (Table~\ref{tab:robotwin_easy}a), where the margin over the strongest baseline widens on tasks demanding precise placement (e.g., 87\% vs.\ 58\% on Place Shoe).

\noindent\textbf{(2) Fine-grained manipulation.} On ManiSkill2 we reach \textbf{55.2\%} on average, ahead of ACT+PonderV2 (42.4\%), DP+PointNet (34.0\%), DP+SPUNet (33.4\%), and DP3 (23.0\%). The gain concentrates where precise spatial understanding is needed: on the ``PegInsertion'' Align and Insert subtasks the strongest baseline reaches only 23\% and 2\%, whereas we hold 75\% and 21\%.

\noindent\textbf{(3) Multi-task learning.} Training one language-conditioned policy over the whole Meta-World ``Hard'' suite, we reach \textbf{60.7\%}, above DP3 (37.5\%), 3D FlowMatch Actor \citep{gkanatsios20253dflowmatchactorunified} (43.3\%), and ManiFlow (51.7\%). The margin persists in this shared-policy regime, so the architecture scales beyond one policy per task.

\subsection{Ablation Studies}
\label{sec:ablation}

\begin{table}[t]
\centering
\small
\resizebox{\columnwidth}{!}{%
\begin{tabular}{l ccccc c}
\toprule
Method & \makecell{Place\\Shoe} & \makecell{Move\\Card} & \makecell{Beat\\Hammer} & \makecell{Turn\\Switch} & \makecell{Open\\Micro.} & \textbf{Avg.} \\
\midrule
Pretrained (Global Cond.)       & 60 & 76 & 86 & 28 & 98 & 69.6 \\
Pretrained (Dense Cond.)        & \textbf{92} & 70 & \textbf{94} & 54 & 99 & 81.8 \\
From-Scratch (Dense Cond.) & 60 & 58 & 90 & 55 & 98 & 72.2 \\
\textbf{(Ours)} Pretr.\ (Dense) + EE (1:1) & 87 & \textbf{85} & 91 & \textbf{56} & \textbf{100} & \textbf{83.8} \\
\midrule
\multicolumn{7}{l}{\emph{Sensitivity to Joint:EE loss weight (dense conditioning + aux.\ EE)}} \\
\ \ Joint:EE = 1:0    & \textbf{92} & 70 & 94 & 54 & 100 & 82.0 \\
\ \ Joint:EE = 1:0.1  & 89 & \textbf{85} & \textbf{97} & 54 & 100 & \textbf{85.0} \\
\ \ Joint:EE = 1:0.5  & 85 & 82 & 94 & 51 & 100 & 82.4 \\
\bottomrule
\end{tabular}%
}
\caption{Component ablation and auxiliary-EE loss-weight sensitivity (\%) over five RoboTwin 2.0 tasks (pretrained PointSAM encoder unless noted). EE supervision improves the average across a range of weights---not cherry-picked. ``Open Microwave'' is near-ceiling for all variants; conclusions are unchanged without it.}
\label{tab:ablation_main}
\end{table}

Table~\ref{tab:ablation_main} isolates the three design choices of Section~\ref{sec:architecture}; training curves in Appendices~\ref{app:pretrain} and~\ref{app:dense_conditioning} confirm the same trends.
\begin{itemize}\itemsep2pt \parskip0pt
    \item \textbf{Dense conditioning.} Replacing global with attention-based dense conditioning is the single largest gain (69.6\% $\to$ 81.8\%) and stabilizes training, verifying a design shared by recent 3D policies. It helps most where fine spatial detail matters: \textit{Place Shoe} rises from 60\% to 92\%.
    \item \textbf{Encoder pre-training.} Initializing from our segmentation-pre-trained encoder beats scratch by 9.6 points (81.8\% vs.\ 72.2\%), consistent with Table~\ref{tab:pretrain_tasks}.
    \item \textbf{Auxiliary EE prediction.} The end-effector objective improves the average further, and the gain is not an artifact of one weight: every non-zero Joint:EE ratio beats the 1:0 baseline, peaking at 85.0\% for 1:0.1.
\end{itemize}

\subsection{Real-World Experiments}
\label{sec:real_exp}

\begin{table}[t]
\centering
\small
\resizebox{\columnwidth}{!}{%
\begin{tabular}{l c ccc c c c}
\toprule
& Long-Horizon & \multicolumn{3}{c}{Pick \& Place} & Articulated & Soft Body & \\
\cmidrule(lr){2-2} \cmidrule(lr){3-5} \cmidrule(lr){6-6} \cmidrule(lr){7-7}
Method & \makecell{Stack Three\\Blocks} & \makecell{Place\\Cup} & \makecell{Fit\\Banana} & \makecell{Place\\Kettle} & \makecell{Open\\Drawer} & \makecell{Fold\\Towel} & \textbf{Avg.} \\
\midrule
DP3           & 2  & 12 & 42 & 40 & 28 & 54 & 29.7 \\
DP            & \textbf{12} & 58 & 48 & 64 & 46 & 36 & 44.0 \\
\textbf{Ours} & \textbf{12} & \textbf{82} & \textbf{64} & \textbf{76} & \textbf{64} & \textbf{66} & \textbf{60.7} \\
\bottomrule
\end{tabular}%
}
\caption{Success rates (\%) on six real-world tasks spanning four task types (50 trials per entry).}
\label{tab:realworld_6task}
\end{table}

\noindent\textbf{Setup.}
We evaluate on an xArm6 with two Intel RealSense D435 RGB-D cameras, one eye-to-hand and one eye-in-hand (Figure~\ref{fig:exp-overall}a,b). Six tasks span four types: pick-and-place (\textit{Place Cup}, \textit{Fit Banana on Plate}, and the pose-constrained \textit{Place Kettle on Stove}), long-horizon (\textit{Stack Three Blocks}), articulated-object (\textit{Open Drawer}), and soft-body manipulation (\textit{Fold Towel}), with object poses randomized at evaluation; task definitions and success criteria are in Appendix~\ref{app:real_world}.
We collect 50 teleoperated demonstrations per task; depth is enhanced with the Camera Depth Model \citep{liu2025manipulation}, and multi-view clouds are fused in the base frame, cropped, and subsampled to 8192 points. We compare our method (ViT-small, following the scaling analysis) against DP3, DP, ManiFlow, and $\pi_0$ on identical real-time fused input, 50 trials per entry. Beyond overall success, we probe two robustness factors---\textbf{(1)} the number of camera views and \textbf{(2)} external lighting interference---on one representative task per type: \textit{Place Kettle} (pose-constrained pick-and-place), \textit{Open Drawer} (articulated), and \textit{Fold Towel} (soft-body).

\noindent\textbf{Overall Performance.} Table~\ref{tab:realworld_6task} shows we lead on every task type, raising the average from 44.0\% (DP) and 29.7\% (DP3) to \textbf{60.7\%}. Table~\ref{tab:single_view} reports the two robustness studies, adding $\pi_0$ and ManiFlow as further baselines.

\noindent\textbf{(1) Number of camera views.} DP3 and ManiFlow use a single eye-to-hand camera, \textbf{underutilizing the capacity of point clouds to fuse multi-view observations}; few prior 3D baselines explore multiple cameras. We fuse the views via hand-eye calibration and evaluate all methods under one and two views.
All methods improve with a second camera, and ours leads under both. With one view, policies show larger errors on tasks needing fine positional control, such as grasping the kettle handle, and often freeze mid-execution; dense feature conditioning both exploits multi-view geometry better and stays robust when viewpoints are limited.

\noindent\textbf{(2) External interference.} Point cloud colors are necessary for some tasks, such as stacking a ``blue'' block on a ``red'' one, yet are easily disturbed by lighting. We therefore add disco-light illumination that dynamically changes scene colors. All methods degrade relative to normal two-view lighting, but ours keeps the highest success rate and the smallest drop.


\begin{table}[t]
\centering
\small
\setlength{\tabcolsep}{3.2pt}
\begin{tabular}{ll cccc c}
\toprule
Task & Views & DP & $\pi_0$ & DP3 & ManiFlow & \textbf{Ours} \\
\midrule
\multirow{3}{*}{\textit{Place Kettle}}
 & Single & 28 & -- & 18 & 20 & \textbf{40} \\
 & Two    & 64 & 46 & 40 & 36 & \textbf{76} \\
 & Disco  & 44 & 40 & 32 & 32 & \textbf{58} \\
\midrule
\multirow{3}{*}{\textit{Open Drawer}}
 & Single & 36 & -- & 12 & 40 & \textbf{52} \\
 & Two    & 46 & 48 & 28 & 46 & \textbf{64} \\
 & Disco  & 34 & 38 & 22 & 36 & \textbf{56} \\
\midrule
\multirow{3}{*}{\textit{Fold Towel}}
 & Single & 28 & -- & 48 & 54 & \textbf{62} \\
 & Two    & 36 & 62 & 54 & 60 & \textbf{66} \\
 & Disco  & 32 & 54 & 38 & 54 & \textbf{62} \\
\midrule
\multirow{3}{*}{\textbf{Average}}
 & Single & 30.7 & -- & 26.0 & 38.0 & \textbf{51.3} \\
 & Two    & 48.7 & 52.0 & 40.7 & 47.3 & \textbf{68.7} \\
 & Disco  & 36.7 & 44.0 & 30.7 & 40.7 & \textbf{58.7} \\
\bottomrule
\end{tabular}
\caption{Real-world robustness study (\%) on one representative task per task type. \textbf{Single}: one camera view; \textbf{Two}: two fused views; both under normal lighting. \textbf{Disco}: two views under disco lighting. $\pi_0$ is originally pre-trained with two views, so its single-view results are not reported.}
\label{tab:single_view}
\end{table}


\section{Conclusion}
We traced the reported instabilities of 3D policy learning to two overlooked details: an EMA bug in the popular Diffusion Policy codebase that cripples any BN-based policy---once fixed, normalization schemes perform comparably, overturning DP3's conclusion that BN must be avoided---and the omission of 3D data augmentation, a major source of overfitting. On these foundations, our study shows that 3D segmentation pre-training transfers reliably, the optimal encoder size scales with point cloud density, deeper action decoders are not necessarily better, and multi-camera fusion substantially improves real-world performance. Our approach outperforms state-of-the-art baselines across three simulation benchmarks and six real-world tasks with the lowest latency, offering a reliable foundation for future work.

\bibliography{main}

\clearpage
\appendix
\setcounter{figure}{0}
\setcounter{table}{0}
\setcounter{equation}{0}
\renewcommand{\thefigure}{\Alph{figure}}
\renewcommand{\thetable}{\Alph{table}}
\renewcommand{\theequation}{A\arabic{equation}}

\setcounter{topnumber}{3}
\setcounter{dbltopnumber}{2}
\setcounter{bottomnumber}{2}
\setcounter{totalnumber}{5}
\renewcommand{\topfraction}{0.92}
\renewcommand{\dbltopfraction}{0.92}
\renewcommand{\bottomfraction}{0.70}
\renewcommand{\textfraction}{0.08}
\renewcommand{\floatpagefraction}{0.80}
\renewcommand{\dblfloatpagefraction}{0.80}
\section{Training Stability and Data Augmentation}
\label{app:augmentation}

\subsection{Correction to EMA Buffer Updates}
\label{app:ema_fix}

The failure of Batch Normalization described in the main paper originates in the rollout copy maintained by the Exponential Moving Average (EMA) updater. The affected implementation updates trainable parameters but does not correctly maintain BatchNorm's non-trainable state. Consequently, the rollout policy can use stale \texttt{running\_mean}, \texttt{running\_var}, and \texttt{num\_batches\_tracked} buffers even when its weights have been updated. Our corrected updater performs the parameter EMA update and explicitly synchronizes these three buffers from the online policy to the rollout policy. LayerNorm and RMSNorm do not carry such running statistics and are therefore unaffected. The normalization comparisons in the main paper otherwise use the same data and training pipeline.

\subsection{Augmentation Hyperparameters}

The augmentation pipeline ablated in the main paper uses the following settings:
\begin{itemize}
    \item \textbf{FPS randomization.} The point order supplied to farthest point sampling is independently reshuffled at every training step. Consequently, the selected subset can vary across epochs for the same frame.
    \item \textbf{Point-cloud color jitter.} Brightness is sampled from $[-0.125, 0.125]$, while contrast and saturation are sampled from $[0.5, 1.5]$. The resulting transform is applied to the RGB channels of the point cloud.
    \item \textbf{Point-cloud noise.} Zero-mean Gaussian noise with standard deviation $\sigma=0.002$ is added to point coordinates.
    \item \textbf{Robot-state noise.} Zero-mean Gaussian noise with standard deviation $\sigma=0.0002$ is added to the proprioceptive state vector.
    \item \textbf{Point dropout.} For each training sample, a dropout ratio is drawn uniformly from $(0,0.8)$ and the corresponding fraction of points is removed.
\end{itemize}
The ``All Aug.'' configuration in the main paper applies all five transforms. The FP3-style comparison uses point dropout, whereas the ManiFlow-style comparison follows its reported image/point-cloud augmentation combination; both are evaluated inside our otherwise unchanged training pipeline.

\subsection{Learning Behavior}

Figure~\ref{fig:data-aug-supp} provides the checkpoint-wise evidence referred to in the main paper. Augmentation produces higher and more stable rollout success, with the largest sustained separation on \textit{Place Shoe}.

\begin{center}
    \begin{minipage}{\columnwidth}
        \centering
        \includegraphics[width=\columnwidth]{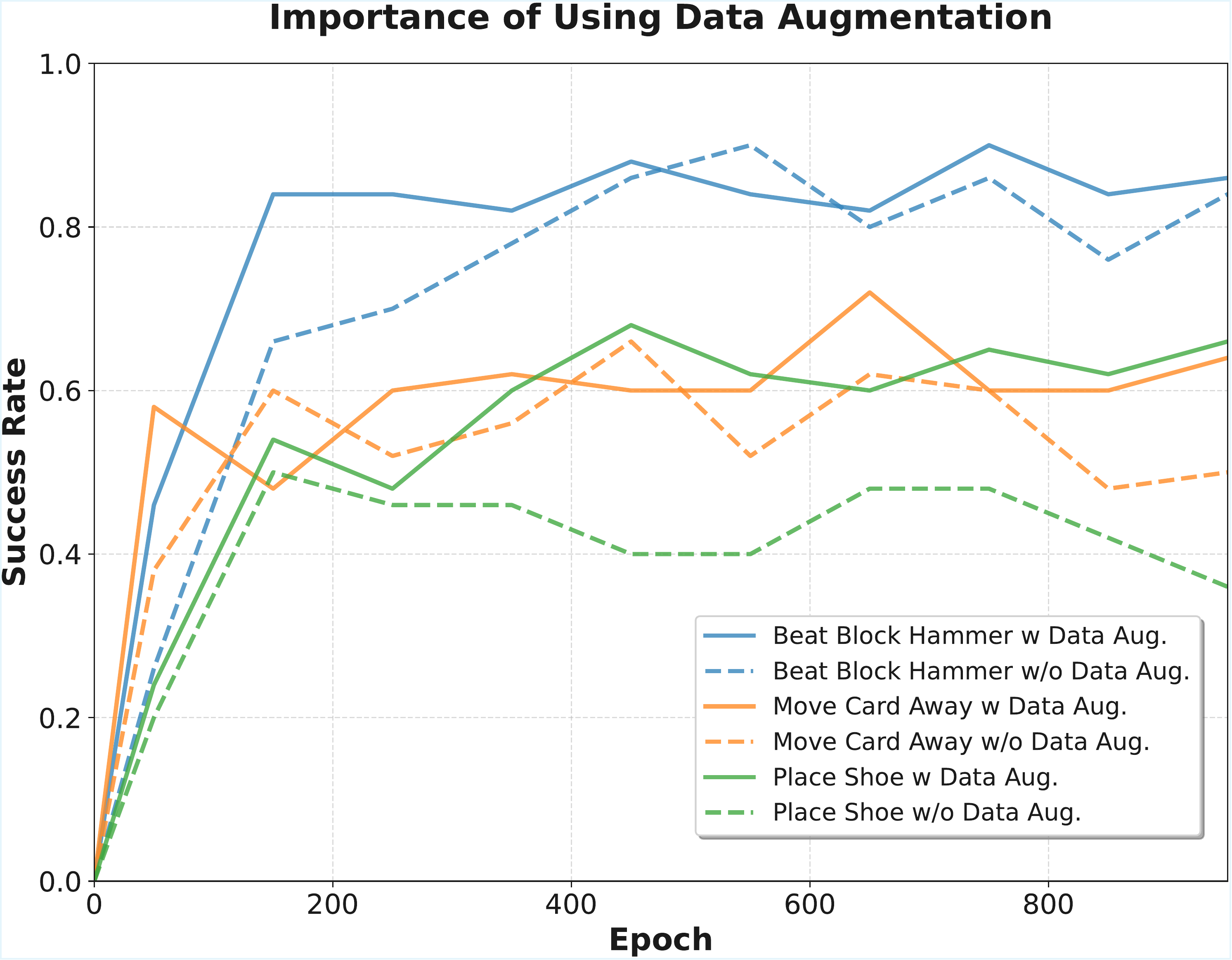}
        \captionof{figure}{\textbf{Effect of data augmentation during training.} Checkpoint success rates on three RoboTwin 2.0 tasks with the full augmentation pipeline (solid) and without augmentation (dashed).}
        \label{fig:data-aug-supp}
    \end{minipage}
\end{center}

\FloatBarrier
\section{Encoder Pre-Training and Scaling}
\label{app:pretrain}

\subsection{Pre-Training Data}

Our pre-training corpus is anchored by indoor scene scans from ScanNet \citep{dai2017scannet} and ARKitScenes \citep{baruch2021arkitscenes}, and is augmented with the synthetic PartNeXt dataset \citep{wangpartnext}, which contains part-level annotations at multiple granularities. Because representing a complete indoor scan with only 1024 or 8192 points would be excessively sparse, we partition large scenes into local blocks before sampling. This preserves local geometric detail at the input resolutions used by the policies. PartNeXt complements the scene data with detailed object and part geometry that is relevant to robotic interaction.

\subsection{Pre-Training Details}

We follow the Point-SAM \citep{zhou2024point} segmentation pre-training protocol with two changes. First, we replace its Voronoi patchification with $k$-nearest-neighbor grouping, which better matches our sampled point-cloud inputs. Second, instead of the original Uni3D \citep{zhou2023uni3d} ViT-large backbone, we use ViT-tiny, ViT-small, and ViT-base variants and scale the embedding dimensions and related hyperparameters accordingly. Each encoder is pre-trained at the point count used by its downstream policy (1024 or 8192 points). Segmentation pre-training runs for 40 epochs on 8 NVIDIA H20 GPUs, after which the resulting weights initialize the policy encoder.

\subsection{Pre-Training Learning Curves}

Figure~\ref{fig:pretrain-supp} shows the additional learning curves referenced by the component ablation in the main paper. The pre-trained encoder improves both early learning and final success, particularly on \textit{Place Shoe}.

\begin{center}
    \begin{minipage}{\columnwidth}
        \centering
        \includegraphics[width=\columnwidth]{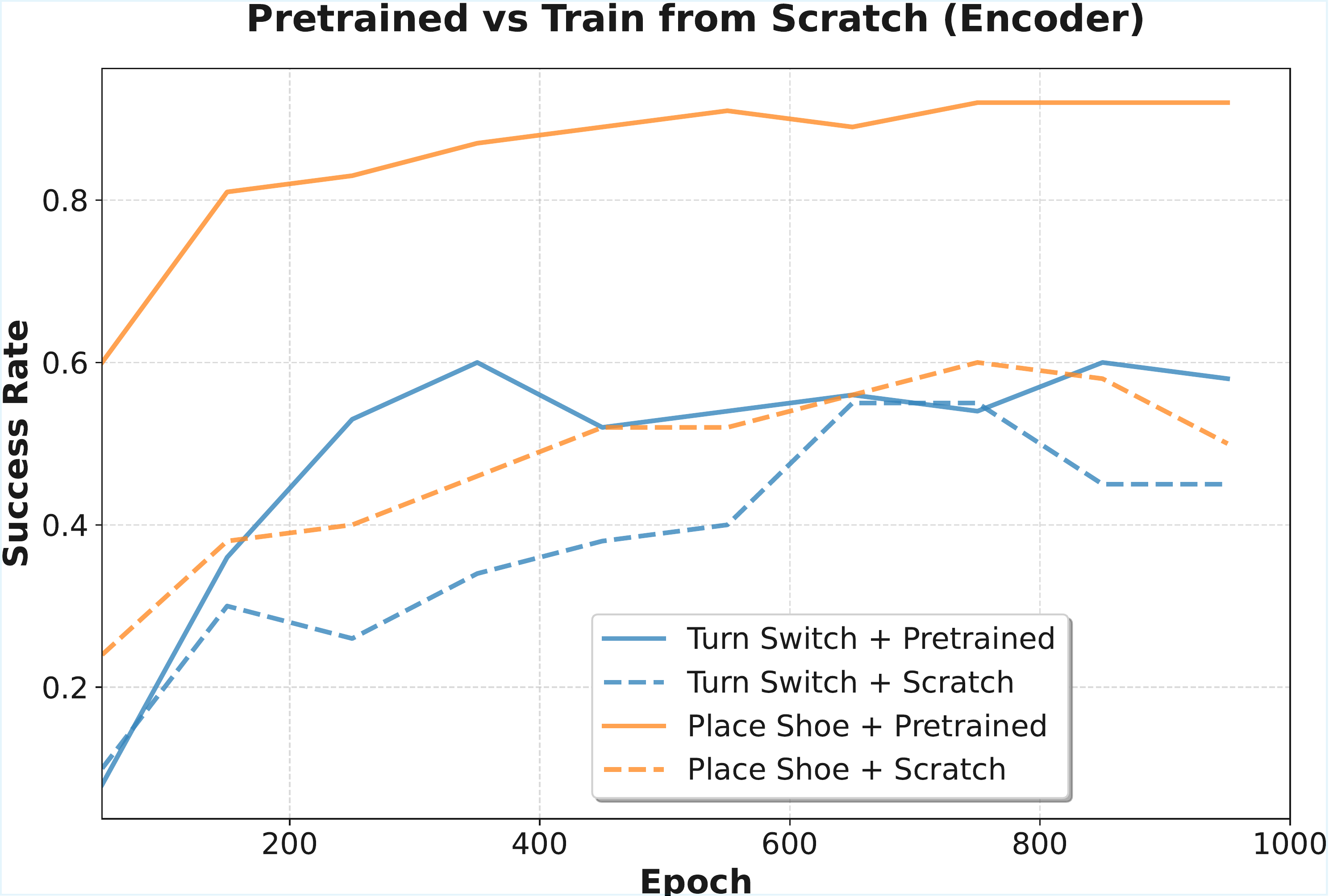}
        \captionof{figure}{\textbf{Segmentation pre-training versus training from scratch.} Checkpoint success rates on two RoboTwin 2.0 tasks. Solid lines use the pre-trained encoder; dashed lines use the same encoder architecture initialized from scratch.}
        \label{fig:pretrain-supp}
    \end{minipage}
\end{center}

\FloatBarrier
\subsection{Per-Task Encoder Scaling Results}
\label{app:encoder_scaling}

Table~\ref{tab:encoder-size-pertask} expands the 50-demonstration point-density comparison from the main paper into its five constituent tasks. ViT-tiny is strongest at 1024 points, whereas the higher-density 8192-point input favors larger encoders on average.

\begin{center}
    \begin{minipage}{\columnwidth}
        \captionof{table}{\textbf{Per-task encoder scaling at 50 demonstrations.} Success rates (\%) for ViT-tiny (53\,MB), ViT-small (115\,MB), and ViT-base (366\,MB). Bold marks the best encoder within each input resolution.}
        \label{tab:encoder-size-pertask}
        \vspace{3pt}
        \centering
        \small
        \setlength{\tabcolsep}{3.5pt}
        \begin{tabular}{l ccc}
            \toprule
            \multicolumn{4}{c}{(a) 1024 points (RoboTwin 2.0 default)} \\
            \cmidrule(lr){1-4}
            Task & Tiny & Small & Base \\
            \midrule
            Move Playingcard  & 85 & \textbf{92} & 83 \\
            Turn Switch       & \textbf{56} & 55 & 50 \\
            Place Shoe        & \textbf{87} & 68 & 70 \\
            Open Microwave    & \textbf{100} & 93 & 77 \\
            Beat Block Hammer & \textbf{91} & 76 & 85 \\
            \midrule
            \textbf{Average}  & \textbf{83.8} & 76.8 & 73.0 \\
            \bottomrule
        \end{tabular}

        \vspace{5pt}

        \begin{tabular}{l ccc}
            \toprule
            \multicolumn{4}{c}{(b) 8192 points (real-world density)} \\
            \cmidrule(lr){1-4}
            Task & Tiny & Small & Base \\
            \midrule
            Move Playingcard  & 65 & \textbf{87} & 72 \\
            Turn Switch       & 47 & 56 & \textbf{64} \\
            Place Shoe        & 75 & 79 & \textbf{85} \\
            Open Microwave    & 79 & \textbf{98} & 92 \\
            Beat Block Hammer & 74 & \textbf{89} & 88 \\
            \midrule
            \textbf{Average}  & 68.0 & \textbf{81.8} & 80.2 \\
            \bottomrule
        \end{tabular}
    \end{minipage}
\end{center}

For completeness, Table~\ref{tab:demo-scaling} reproduces the demonstration-count comparison used to establish that the preferred 1024-point encoder size remains stable as the amount of policy data increases.

\begin{center}
    \begin{minipage}{\columnwidth}
        \captionof{table}{\textbf{Scaling with policy demonstrations.} Average success rate (\%) over five RoboTwin 2.0 ``Easy'' tasks at 1024 points.}
        \label{tab:demo-scaling}
        \vspace{3pt}
        \centering
        \small
        \setlength{\tabcolsep}{4pt}
        \begin{tabular}{l ccc}
            \toprule
            Encoder & 50 demos & 100 demos & 200 demos \\
            \midrule
            ViT-tiny  (53\,MB)  & \textbf{83.8} & \textbf{95.3} & \textbf{97.8} \\
            ViT-small (115\,MB) & 76.8 & 89.0 & 92.4 \\
            ViT-base  (366\,MB) & 73.0 & 86.5 & 88.2 \\
            \bottomrule
        \end{tabular}
    \end{minipage}
\end{center}

\FloatBarrier
\section{Dense Action Conditioning}
\label{app:dense_conditioning}

\subsection{Decoder Architecture Comparison}

Figure~\ref{fig:decoder-comparison} is the schematic referenced by the dense-conditioning discussion in the main paper. DP3 compresses the observation into a global vector and injects it through FiLM. Our decoder instead retains geometric and proprioceptive tokens and exposes them to every denoising block through cross-attention. The self-attention mask also shows the joint-action and auxiliary end-effector token streams.

\begin{center}
    \begin{minipage}{\columnwidth}
        \centering
        \includegraphics[width=0.96\columnwidth,trim=0 0 550bp 0,clip]{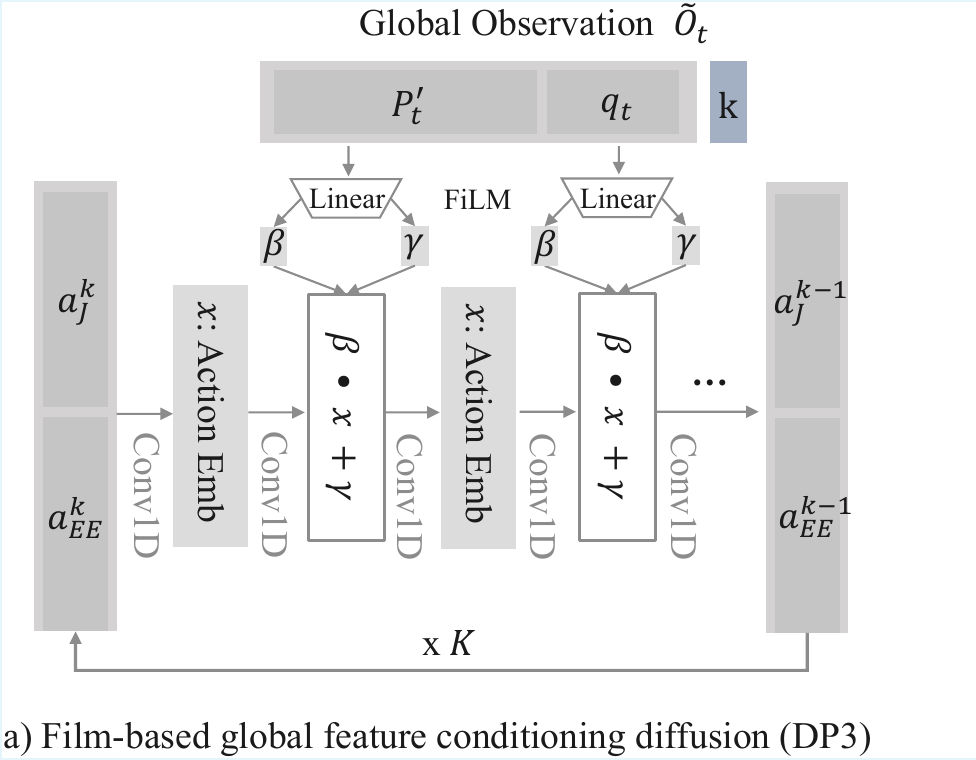}\\[5pt]
        \includegraphics[width=0.96\columnwidth,trim=470bp 0 0 0,clip]{figures/figure-1-1-131-new_cropped.pdf}
        \captionof{figure}{\textbf{Global versus dense conditioning.} (a) DP3 uses a pooled observation feature to modulate the convolutional denoiser through FiLM. (b) Our DiT retains spatially resolved geometric features and conditions action denoising through cross-attention; the left inset shows self-attention between joint-space and auxiliary end-effector action tokens.}
        \label{fig:decoder-comparison}
    \end{minipage}
\end{center}

\subsection{Dense-Conditioning Learning Curves}

The checkpoint curves in Figure~\ref{fig:dense-conditioning-supp} complement the aggregate component ablation in the main paper. Dense conditioning has the clearest advantage on \textit{Turn Switch}, where the global representation becomes unstable late in training.

\begin{center}
    \begin{minipage}{\columnwidth}
        \centering
        \includegraphics[width=\columnwidth]{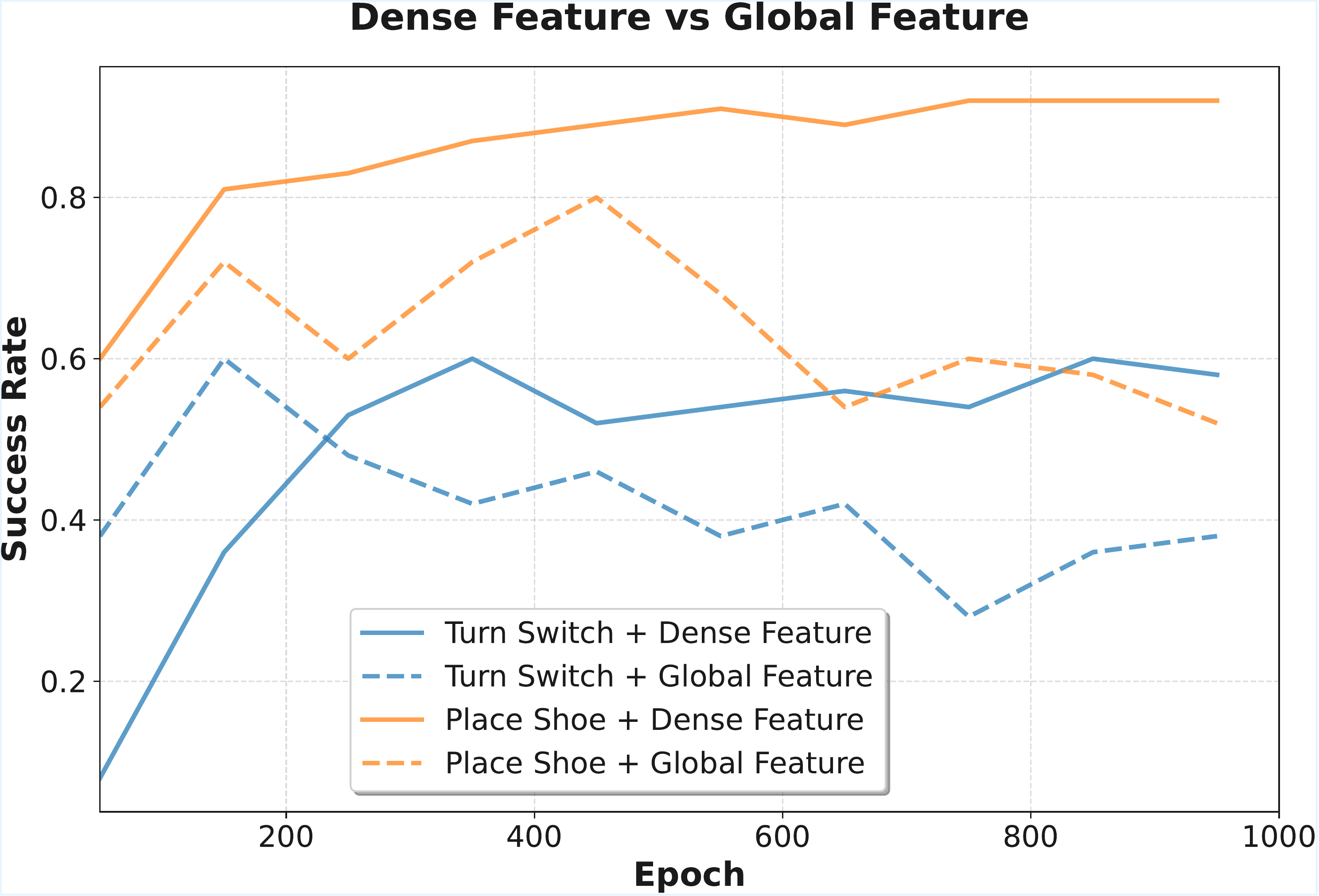}
        \captionof{figure}{\textbf{Dense versus global conditioning.} Checkpoint success rates on two RoboTwin 2.0 tasks. Solid lines use attention-based dense conditioning; dashed lines use a pooled global feature.}
        \label{fig:dense-conditioning-supp}
    \end{minipage}
\end{center}

\FloatBarrier
\section{Simulation Experiments}
\label{app:simulation}

\subsection{Common Protocol}

RoboTwin 2.0 \citep{chen2025robotwin} experiments use 50 demonstrations per task, 1024-point inputs, and one policy per task. The ``Easy'' suite contains clean scenes, whereas the ``Hard'' suite adds distractors and variations in appearance, geometry, object pose, table height, and illumination. ManiSkill2 \citep{gu2023maniskill2} uses 1000 demonstrations per task and 1024-point inputs. Success rate is the evaluation metric throughout.

\begin{figure*}[!t]
    \centering
    \includegraphics[width=\textwidth]{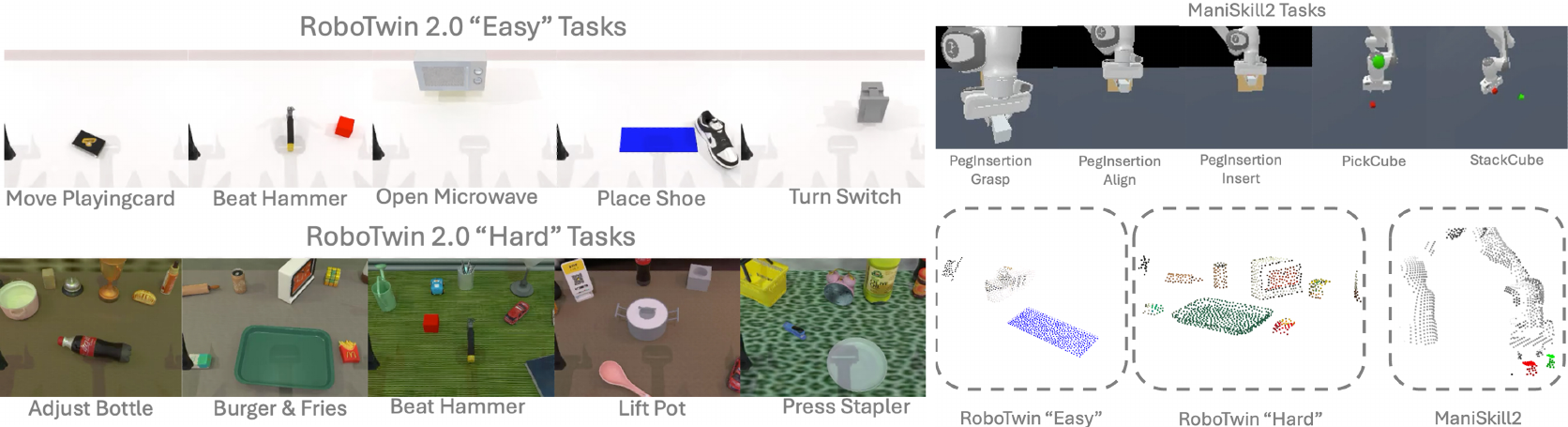}
    \caption{\textbf{Simulation tasks and point-cloud inputs.} The top and middle rows show the selected RoboTwin 2.0 ``Easy,'' RoboTwin 2.0 ``Hard,'' and ManiSkill2 tasks. The lower row shows representative cropped and downsampled point clouds from each benchmark family.}
    \label{fig:simulation-tasks}
\end{figure*}

Figure~\ref{fig:simulation-tasks} shows the selected tasks and representative policy inputs. The lower row illustrates the post-cropping, 1024-point observations rather than the raw simulator render.

\subsection{RoboTwin 2.0 Baseline Configurations}

The main 3D baselines use the following configurations:
\begin{itemize}
    \item \textbf{DP3} \citep{ze20243d} is trained for 3000 epochs with batch size 256, planning horizon 8, and 1024 input points after background and tabletop segmentation.
    \item \textbf{iDP3} \citep{ze2024idp3} follows its published improved-DP3 architecture and official task configuration at the common 1024-point input resolution.
    \item \textbf{ManiFlow} \citep{yan2025maniflowgeneralrobotmanipulation} uses a 12-layer DiTX decoder with 8 attention heads and embedding dimension 768. Its DP3Encoder processes 128 downsampled points. Training runs for 1000 epochs with batch size 384 on one 24\,GB NVIDIA RTX 4090.
    \item \textbf{Ours} is trained for 1000 epochs with batch size 256, planning horizon 16, and 1024 input points.
\end{itemize}

\noindent\textbf{Additional ``Easy'' baselines.}
\begin{itemize}
    \item \textbf{Diffusion Policy (DP)} \citep{chi2024diffusionpolicyvisuomotorpolicy} is trained for 600 epochs with batch size 128 and planning horizon 8.
    \item \textbf{RDT} \citep{liu2025rdt1bdiffusionfoundationmodel} is pre-trained for 100{,}000 steps with batch size 16 per GPU on 8 GPUs, then fine-tuned per task for 10{,}000 steps with batch size 16 per GPU on 4 GPUs.
    \item \textbf{Spatial Forcing} \citep{li2025spatialforcingimplicitspatial} uses $\pi_0$ with PaliGemma as its base model. We train its LoRA adaptation for 30{,}000 iterations on one NVIDIA H100 and otherwise follow the official settings.
    \item \textbf{DP+eVGGT} \citep{vuong2025improvingroboticmanipulationefficient} uses the distilled RoboTwin 2.0 VGGT encoder as a frozen backbone in a UNet-based diffusion policy. The policy has horizon 8, 3 observation steps, 6 action steps, and UNet widths $[256,512,1024]$; it is trained for 100 epochs with batch size 48.
\end{itemize}

\noindent\textbf{Additional ``Hard'' baselines.}
\begin{itemize}
    \item \textbf{DP+eVGGT} uses the same model configuration as above and is trained for 60--80 epochs with batch size 48.
    \item \textbf{ACT+eVGGT} \citep{vuong2025improvingroboticmanipulationefficient} uses chunk size 50 and batch size 8, is trained for 6000 epochs on one GPU, and applies temporal aggregation during deployment.
    \item \textbf{RDT} uses the pre-training and fine-tuning schedule given above.
\end{itemize}

\subsection{ManiSkill2 Baseline Configurations}

Observations are cropped to the workspace and downsampled to 1024 points using farthest point sampling. For \textit{PickCube}, the goal is rendered as a green sphere so it is observable in the point cloud.
\begin{itemize}
    \item \textbf{DP+SPUNet} \citep{zhu2024point} uses SPUNet34 in the official UNet-based Diffusion Policy. It is trained for 1800 epochs with batch size 32, prediction horizon 16, 8 action steps, and 2 observation steps. The UNet widths are $[512,1024,2048]$ and the diffusion-step embedding dimension is 128.
    \item \textbf{DP+PointNet} \citep{zhu2024point} uses the standard PointNet encoder with Diffusion Policy. It is trained for 1800 epochs with AdamW (weight decay 0.05) and a OneCycle schedule, using prediction horizon 16.
    \item \textbf{ACT+PonderV2} \citep{zhu2024point} combines ACT with the SPUNet34-based PonderV2 backbone. ACT has 4 encoder layers, 7 decoder layers, hidden dimension 512, latent dimension 32, and KL weight 10.0. It is trained for 500 epochs with chunk size 100 and uses temporal ensembling with decay $k=0.01$.
    \item \textbf{DP3} is trained for 1000 epochs with batch size 2048, planning horizon 8, 1024 points, and no point-cloud RGB.
    \item \textbf{Ours} is trained for 1000 epochs with batch size 2048, planning horizon 16, 1024 points, and point-cloud color.
\end{itemize}

\subsection{Per-Task ManiSkill2 Results}
\label{app:maniskill}

Table~\ref{tab:maniskill-pertask} is the per-task breakdown of the averages quoted in the main paper. The largest separation occurs in the fine-grained Align and Insert stages of \textit{PegInsertionSide}.

\begin{center}
    \begin{minipage}{\columnwidth}
        \captionof{table}{\textbf{Per-task ManiSkill2 results.} Success rates (\%); \textit{PegInsertionSide} is decomposed into Grasp, Align, and Insert stages.}
        \label{tab:maniskill-pertask}
        \vspace{3pt}
        \centering
        \small
        \setlength{\tabcolsep}{2pt}
        \resizebox{\columnwidth}{!}{%
        \begin{tabular}{l ccccc c}
            \toprule
            Method
            & \makecell{Pick\\Cube}
            & \makecell{Stack\\Cube}
            & \makecell{PegIns.\\Grasp}
            & \makecell{PegIns.\\Align}
            & \makecell{PegIns.\\Insert}
            & \textbf{Avg.} \\
            \midrule
            DP+SPUNet     & 71 & 4  & 82 & 9  & 1 & 33.4 \\
            DP+PointNet   & 70 & 0  & 83 & 16 & 1 & 34.0 \\
            ACT+PonderV2  & 87 & \textbf{35} & 65 & 23 & 2 & 42.4 \\
            DP3           & 57 & 14 & 41 & 3  & 0 & 23.0 \\
            \midrule
            \textbf{Ours} & \textbf{97} & 24 & \textbf{97}
                          & \textbf{75} & \textbf{21} & \textbf{55.2} \\
            \bottomrule
        \end{tabular}%
        }
    \end{minipage}
\end{center}

\subsection{Meta-World Multi-Task Evaluation}
\label{app:metaworld}

Meta-World \citep{yu2019metaworld} evaluates a different scaling axis: a single language-conditioned policy is trained across the complete ``Hard'' suite rather than fitting a separate policy to each task. The comparison includes DP3, 3D FlowMatch Actor \citep{gkanatsios20253dflowmatchactorunified}, ManiFlow, and our method. Table~\ref{tab:metaworld-supp} reports the common-suite averages used in the main paper.

\begin{center}
    \begin{minipage}{\columnwidth}
        \captionof{table}{\textbf{Meta-World ``Hard'' multi-task results.} One language-conditioned policy is shared across tasks.}
        \label{tab:metaworld-supp}
        \vspace{3pt}
        \centering
        \small
        \setlength{\tabcolsep}{5pt}
        \begin{tabular}{l c}
            \toprule
            Method & Success rate (\%) \\
            \midrule
            DP3 & 37.5 \\
            3D FlowMatch Actor & 43.3 \\
            ManiFlow & 51.7 \\
            \textbf{Ours} & \textbf{60.7} \\
            \bottomrule
        \end{tabular}
    \end{minipage}
\end{center}

\FloatBarrier
\section{Real-World Experiments}
\label{app:real_world}

\subsection{Robot and Camera Setup}

Figure~\ref{fig:real-setup} shows the xArm6 platform, gripper, and two Intel RealSense D435 RGB-D cameras. The eye-to-hand camera provides a stable global view, while the eye-in-hand camera supplies close-range geometry. The inset shows the disco-light condition used in the robustness study.

\begin{center}
    \begin{minipage}{\columnwidth}
        \centering
        \includegraphics[width=0.92\columnwidth]{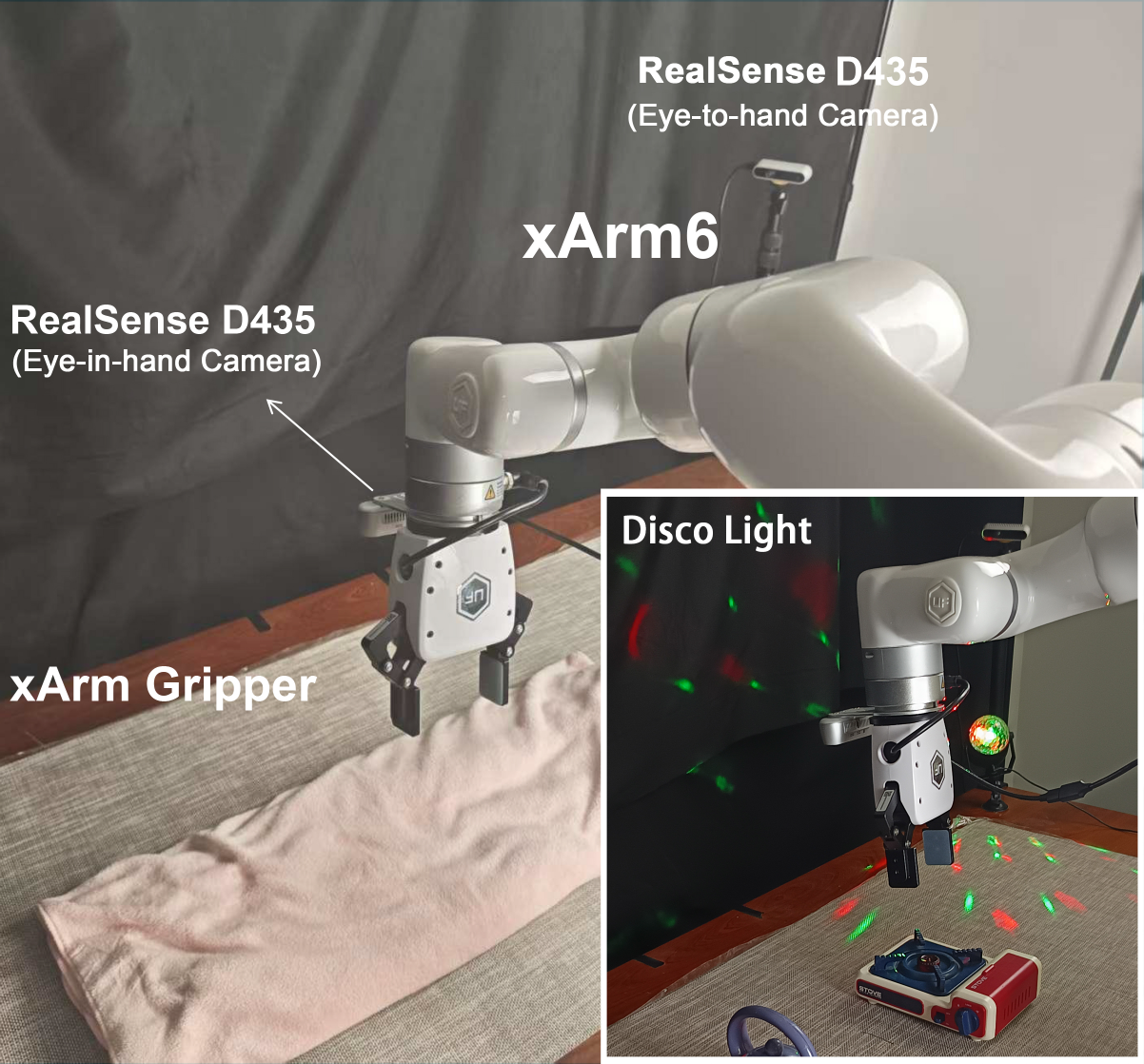}
        \captionof{figure}{\textbf{Real-world setup.} An xArm6 with eye-to-hand and eye-in-hand RealSense D435 cameras. The inset shows the disco-light interference condition.}
        \label{fig:real-setup}
    \end{minipage}
\end{center}

\FloatBarrier
\subsection{Task Definitions and Success Criteria}

We evaluate six tasks spanning four manipulation types:
\begin{itemize}
    \item \textbf{Place Cup.} A vertical cup starts in the right half of a $60\,\mathrm{cm}\times40\,\mathrm{cm}$ foam board, with yaw sampled over $360^\circ$. A trial succeeds when the robot grasps the cup, places it upright in either target hole, releases it, and the cup remains stable. Missing the hole, colliding with the rim, or leaving the cup tipped over counts as failure.
    \item \textbf{Fit Banana on Plate.} The plate is fixed on the same board, centered on its mid-line and shifted 10\,cm to the right. A banana-shaped object starts 10--15\,cm to the left with yaw sampled over $360^\circ$. Success requires the released object to lie fully inside the plate boundary. Contact with the rim or a final pose extending beyond the plate is a failure.
    \item \textbf{Stack Three Blocks.} The board is divided along its long side into three $20\,\mathrm{cm}\times40\,\mathrm{cm}$ cells. One block is placed near the center of each cell with center jitter up to $\pm5$\,cm and nearly fixed orientation. Success requires a stable three-block tower in the central stacking region. A collapsed, incomplete, or partially unmoved stack fails.
    \item \textbf{Place Kettle on Stove.} A kettle starts within a $20\,\mathrm{cm}\times20\,\mathrm{cm}$ region with yaw in $\pm60^\circ$. Success requires grasping the handle, aligning the kettle center with the blue stove target, releasing it, and leaving it stable. A missed handle or offset final placement fails.
    \item \textbf{Open Drawer.} A three-drawer unit is placed within an L-shaped region formed by a $70\,\mathrm{cm}\times20\,\mathrm{cm}$ rectangle and a $20\,\mathrm{cm}\times20\,\mathrm{cm}$ rectangle. Success requires grasping the designated front-facing handle and opening the drawer to at least 50\% of its full extension.
    \item \textbf{Fold Towel.} A towel is placed within a $70\,\mathrm{cm}\times50\,\mathrm{cm}$ region with orientation parallel to the table edge plus a yaw perturbation in $\pm30^\circ$. Success requires grasping and folding the towel so its two sides align. A missed grasp or materially misaligned final fold fails.
\end{itemize}

\subsection{Evaluation and Termination Protocol}

Each method is evaluated for 50 trials per task. A trial runs for at most 60 seconds and is marked unsuccessful if the task criterion is not met before the limit or if the robot becomes stuck without meaningful motion. Within a given one-view, two-view, or disco-light condition, methods use the same camera placement, calibration, workspace, and control setup; each method retains its native input representation.

\subsection{Point-Cloud Processing}
\label{app:pcd}

For each demonstration, we reconstruct one colored point cloud in the robot base frame using the pipeline in Figure~\ref{fig:train-pointcloud}:
\begin{enumerate}
    \item \textbf{RGB-D pre-processing.} RGB and depth maps are undistorted and back-projected with calibrated camera intrinsics. During demonstration processing, depth can first be enhanced by the Camera Depth Model (CDM) \citep{liu2025manipulation}.
    \item \textbf{Coordinate transformation.} The eye-to-hand and eye-in-hand clouds are independently transformed into the robot base frame using camera extrinsics from EasyHeC++ hand-eye calibration \citep{hong2024easyhec++}.
    \item \textbf{Multi-view fusion.} The transformed clouds are concatenated into a colored cloud containing approximately 200{,}000--400{,}000 points.
    \item \textbf{Workspace cropping.} A fixed box removes irrelevant background in the horizontal workspace: $x\in[-0.5,0.82]$ and $y\in[-0.7,0.8]$.
    \item \textbf{Training-time sampling.} The cropped cloud is reduced to 8192 points with farthest point sampling, preserving coverage of manipulable surfaces.
    \item \textbf{Deployment-time sampling.} Point clouds are generated online at approximately 30\,Hz. CDM is disabled for latency; raw RGB-D frames are fused, transformed, cropped, and uniformly sampled to 8192 points.
\end{enumerate}

\begin{center}
    \begin{minipage}{\columnwidth}
        \centering
        \scriptsize
        \begin{tabular}{@{}cc@{}}
            \includegraphics[width=0.46\columnwidth]{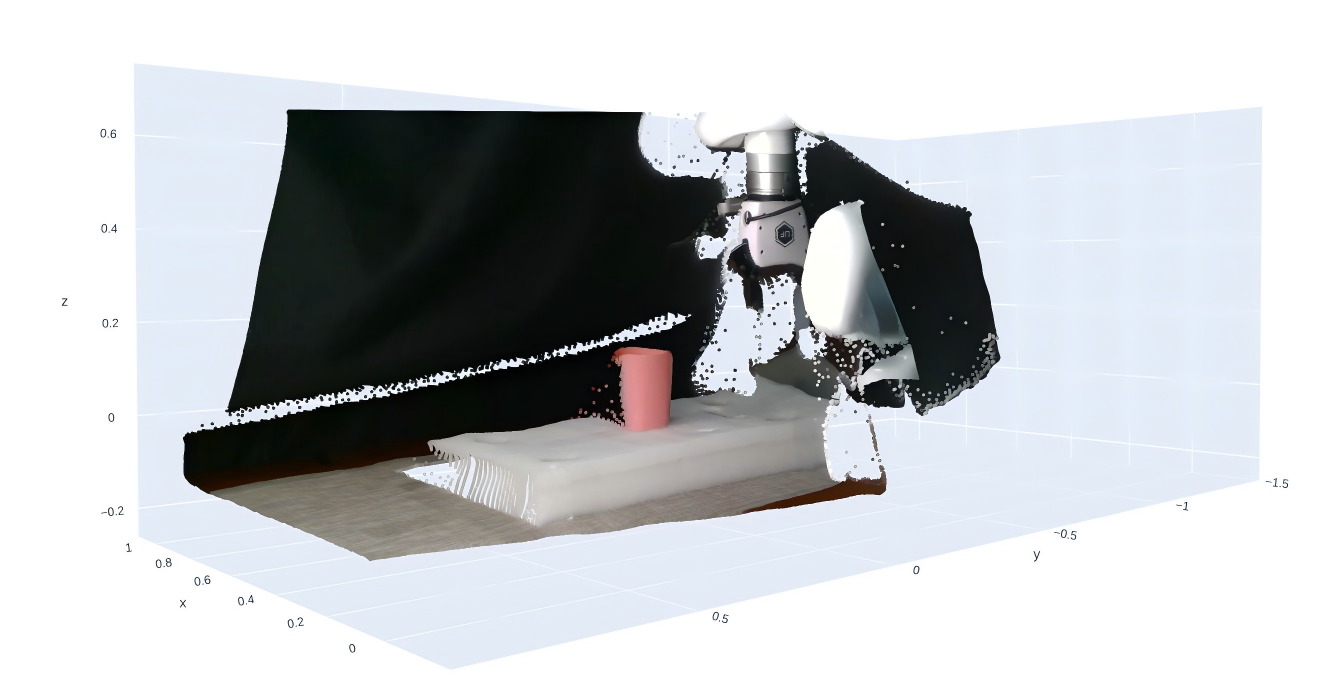} &
            \includegraphics[width=0.46\columnwidth]{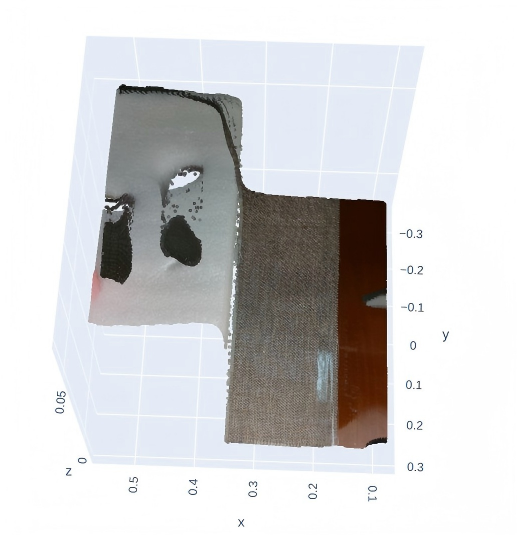} \\
            (a) Eye-to-hand & (b) Eye-in-hand \\[4pt]
            \includegraphics[width=0.46\columnwidth]{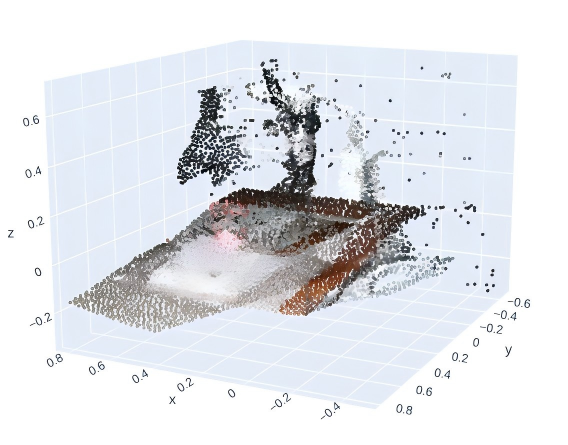} &
            \includegraphics[width=0.46\columnwidth]{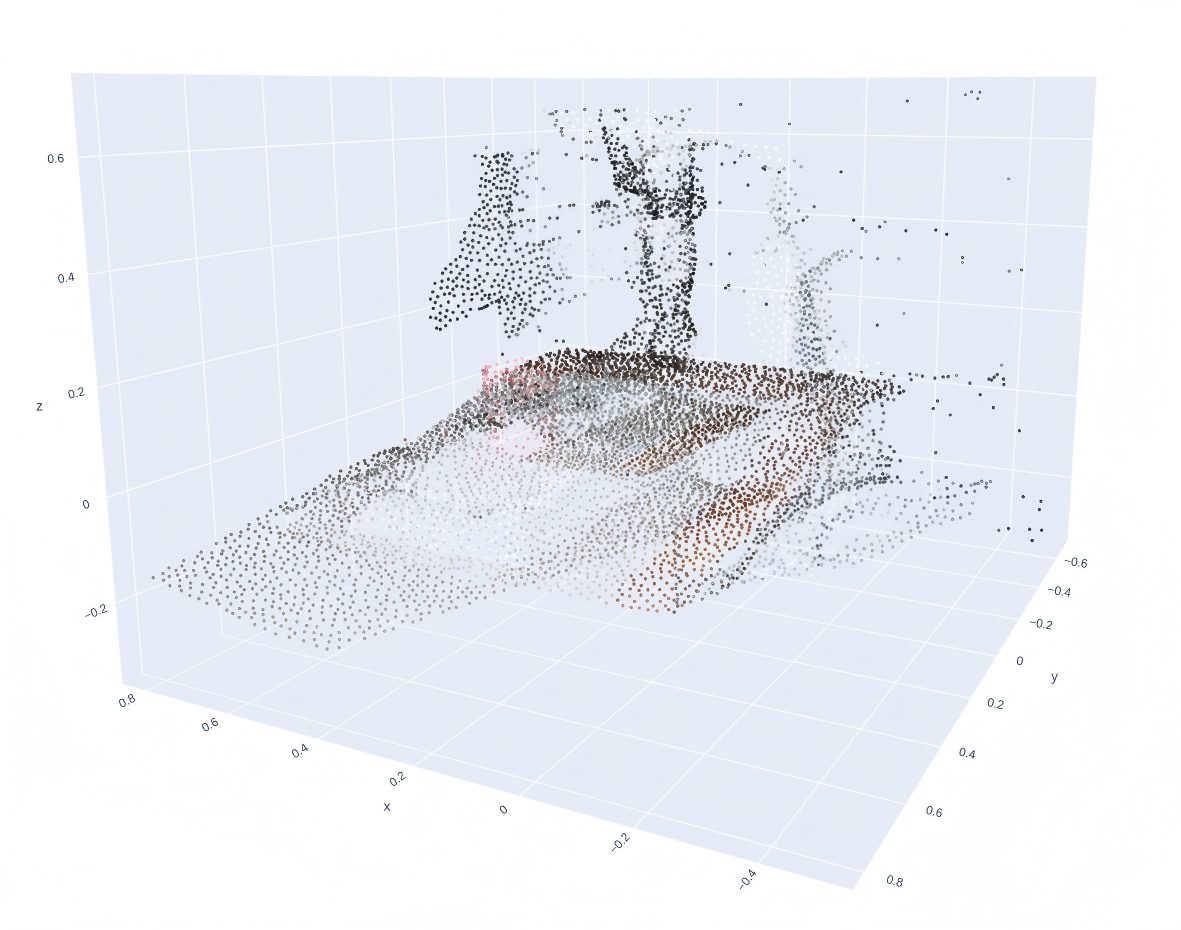} \\
            (c) Merged and cropped & (d) Enlarged training cloud
        \end{tabular}
        \captionof{figure}{\textbf{Training point-cloud pipeline.} (a--b) Each RGB-D view is back-projected and transformed to the robot base frame. (c) The views are merged, workspace-cropped, and sampled to 8192 points. (d) An enlarged view shows the retained local geometry.}
        \label{fig:train-pointcloud}
    \end{minipage}
\end{center}

Figure~\ref{fig:realtime-pointcloud} shows the corresponding fused point cloud used online. It follows the same coordinate and cropping conventions as the training representation, but uses raw depth and uniform sampling to meet the control-time budget.

\begin{center}
    \begin{minipage}{\columnwidth}
        \centering
        \includegraphics[width=0.86\columnwidth]{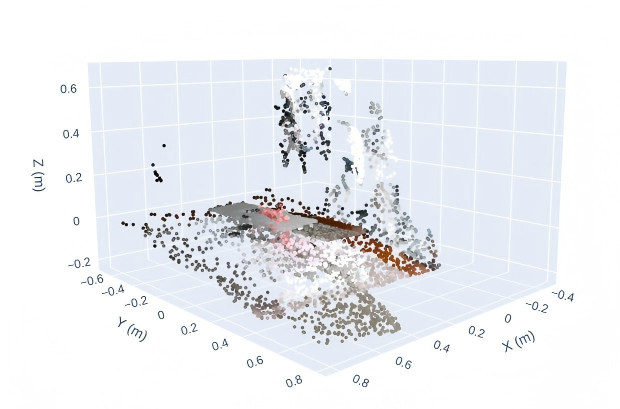}
        \captionof{figure}{\textbf{Real-time fused point cloud.} The deployment pipeline updates the cloud at approximately 30\,Hz and samples 8192 points.}
        \label{fig:realtime-pointcloud}
    \end{minipage}
\end{center}

\subsection{Real-World Training and Compute}

We collect 50 human-teleoperated demonstrations per task. CDM is used offline to construct the training point clouds: refined depth is fused with RGB, transformed to the robot base frame, merged, workspace-cropped, and sampled to 8192 points. CDM is disabled during every real-world evaluation. The real-world policies are trained for 1000 epochs. Policy training and encoder pre-training use a server with 8 NVIDIA H20 GPUs; latency is measured on one NVIDIA RTX 3090. Implementations use PyTorch.

\FloatBarrier
\section{Inference Latency Protocol}
\label{app:latency}

The latency comparison isolates policy execution from sensor pre-processing. All methods are timed on one NVIDIA RTX 3090 under the same real-world sensing condition. Point-cloud policies start from a ready 8192-point fused tensor, while image policies start from the corresponding ready RGB tensors. We discard an initial warm-up period to exclude CUDA kernel autotuning and lazy initialization, then report the mean wall-clock time over 100 complete policy forward passes. A timed pass includes observation encoding, the method's complete iterative denoising or flow-matching schedule, and action decoding. It excludes camera acquisition and input construction, which run outside the policy and are excluded for every method. Each method retains its published number of sampling steps because the sampling schedule is part of the method.

Table~\ref{tab:latency-supp} reproduces the end-to-end measurements for completeness. The encoder/decoder split discussed in the main paper is obtained by timing those two submodules separately under the same protocol; our 4-block DiT decoder takes 35.63\,ms, compared with 98.78\,ms for DP3's UNet decoder.

\begin{center}
    \begin{minipage}{\columnwidth}
        \captionof{table}{\textbf{Inference latency.} Mean end-to-end policy time over 100 post-warm-up runs on one RTX 3090.}
        \label{tab:latency-supp}
        \vspace{3pt}
        \centering
        \small
        \setlength{\tabcolsep}{4pt}
        \begin{tabular}{l c}
            \toprule
            Method & End-to-end latency (ms) \\
            \midrule
            DP & 140.03 \\
            $\pi_0$ & 137.88 \\
            DP3 & 115.42 \\
            ManiFlow & 122.24 \\
            \textbf{Ours} & \textbf{65.26} \\
            \bottomrule
        \end{tabular}
    \end{minipage}
\end{center}

\FloatBarrier

\end{document}